\documentclass[journal]{IEEEtran}

\ifCLASSINFOpdf
  \usepackage[pdftex]{graphicx}
  \graphicspath{{Figures/}}
\else
\fi
\usepackage{amssymb}
\usepackage{multirow}

\usepackage{color}
\usepackage[cmex10]{amsmath}
\usepackage{array}

\usepackage[noend]{algpseudocode}

\usepackage{algorithmicx,algorithm}
\usepackage{array}

\usepackage[noend]{algpseudocode}

\usepackage{algorithmicx,algorithm}

\usepackage{url}
\usepackage[driverfallback=dvipdfm,
CJKbookmarks=true, bookmarksnumbered=true, bookmarksopen=true,
colorlinks=true, %
pdfborder=001, %
citecolor=blue, linkcolor=blue, anchorcolor=red, urlcolor=black
]{hyperref}

\usepackage[tight,footnotesize]{subfigure}

\usepackage{caption}

\usepackage{booktabs}

\usepackage{makecell}

\begin{document}
%
\title{{Weakly-Supervised Temporal Action Localization with Bidirectional Semantic Consistency Constraint}}
%
%
%
\author{
        Guozhang~Li,
        De~Cheng,
        Xinpeng~Ding,
        Nannan~Wang,~\IEEEmembership{Member,~IEEE},
        Jie~Li,
        Xinbo~Gao,~\IEEEmembership{Senior Member,~IEEE}

\thanks{Manuscript received 30 July 2022; revised 21 November 2022 and 5 February 2023; accepted 4 April 2023. This work was supported in part by the National Key Research and Development Program of China under Grant 2018AAA0103202; in part by the National Natural Science Foundation of China under Grants 62036007, U22A2096, 62176195 and 62176198; in part by the Technology Innovation Leading Program of Shaanxi under Grant 2022QFY01-15; in part by Open Research Projects of Zhejiang Lab under Grant 2021KG0AB01; in part by the Fundamental Research Funds for the Central Universities; in part by the Innovation Fund of Xidian University. \emph{(corresponding author: Xinbo Gao and Nannan Wang.)}}
\thanks{G. Li and J. Li are with the State Key Laboratory of Integrated Services Networks, School of Electronic Engineering, Xidian University, Xi'an 710071, Shaanxi, P. R. China (e-mail: liguozhang@stu.xidian.edu.cn and leejie@mail.xidian.edu.cn).} 
\thanks{N. Wang and D. Cheng are with the State Key Laboratory of Integrated Services Networks, School of Telecommunications Engineering, Xidian University, Xi'an 710071, Shaanxi, P. R. China (e-mail: nnwang@xidian.edu.cn and dcheng@xidian.edu.cn).}
\thanks{X. Ding is with the School of Engineering, the Hong Kong University of Science and Technology, Hong Kong, (e-mail: xdingaf@connect.ust.hk).}
\thanks{X. Gao is with the School of Electronic Engineering, Xidian University, Xi'an 710071, China, and also with the Chongqing Key Laboratory of Image Cognition, Chongqing University of Posts and Telecommunications, Chongqing 400065, China (e-mail: xbgao@mail.xidian.edu.cn).}

}

\markboth{$>$ \normalsize{IEEE T}\footnotesize{ransactions on} \normalsize{N}\footnotesize{eural} \normalsize{N}\footnotesize{etworks} \normalsize{A}\footnotesize{nd} \normalsize{L}\footnotesize{earning} \normalsize{S}\footnotesize{ystems} $<$}
{Shell \MakeLowercase{\textit{et al.}}: Bare Demo of IEEEtran.cls for Journals}



\maketitle

\begin{abstract}


Weakly Supervised Temporal Action Localization (WTAL) aims to classify and localize temporal boundaries of actions for the video, given only video-level category labels in the training datasets.
Due to the lack of boundary information during training, existing approaches formulate WTAL as a classification problem,~\emph{i.e.}, generating the temporal class activation map (T-CAM) for localization.
However, with only classification loss, the model would be sub-optimized,~\emph{i.e.}, the action-related scenes are enough to distinguish different class labels.
Regarding other actions in the action-related scene (~\emph{i.e.}, the scene same as positive actions) as co-scene actions, this sub-optimized model would misclassify the co-scene actions as positive actions.
%
%
%
%
To address this misclassification, we propose a simple yet efficient method, named bidirectional semantic consistency constraint (Bi-SCC), to discriminate the positive actions from co-scene actions.
The proposed Bi-SCC firstly adopts {a temporal context augmentation to generate an augmented video that breaks the correlation between positive actions and their co-scene actions in the inter-video};
Then, a semantic consistency constraint (SCC) is used to enforce the predictions of the original video and augmented video to be {consistent}, hence suppressing the co-scene actions.
However, we find that this augmented video would destroy the original temporal context.
Simply applying {the} consistency constraint would affect the completeness of localized positive actions.
Hence, we boost the SCC in a bidirectional way to suppress co-scene actions while ensuring the integrity of positive actions, by cross-supervising the original and augmented videos.
Finally, our proposed Bi-SCC can be applied to current WTAL approaches, and improve their performance.
Experimental results show that our approach outperforms the state-of-the-art methods on THUMOS14 and ActivityNet. {The code is available at https://github.com/lgzlIlIlI/BiSCC.}
\end{abstract}


\IEEEpeerreviewmaketitle

\section{Introduction}
\label{sec:Introduction}

\IEEEPARstart
{T}emporal action localization (TAL) aims to locate the temporal boundaries and predict classes corresponding to the actions of interest in untrimmed videos.
Due to the great application potential in video surveillance \cite{yun2019vision}, event analysis \cite{chang2016semantic}, and others, TAL has attracted lots of attention in recent years.
Currently, owing to the development of deep learning, remarkable progress has been made in the fully supervised scenario \cite{lin2017single,Zeng_2019_ICCV,Lin_2018_ECCV,Lin_2019_ICCV,Long_2019_CVPR,tan2021relaxed,ding2021crnet}.  
However, most recent fully supervised TAL algorithms have become reliant on dense time-consuming and labor-intensive frame-level annotations, limiting the practical application of these methods in real-world scenarios.
To alleviate this limitation, weakly-supervised temporal action localization (WTAL) which focuses on weak supervision setting such as video-level labels \cite{paul2018w,narayan20193c,lee2019background,Shi_2020_CVPR,2020Adversarial,huang2020relational,lee2021weakly,li2021multi} has recently received increasing attention.  
Owing to the lack of precise temporal boundaries, WTAL usually {formulates} this task as the classification problem. In specific, the WTAL method only utilizes the video-level labels to learn the temporal class activation map (T-CAM) for each input video, where the T-CAM indicates the probability that each frame in the video belongs to an action class. Then, the boundaries of action instances in the video can be localized by threshold scores on T-CAM during the test phase.

\begin{figure}[!t]
\begin{center}
    \includegraphics[width=1\linewidth]{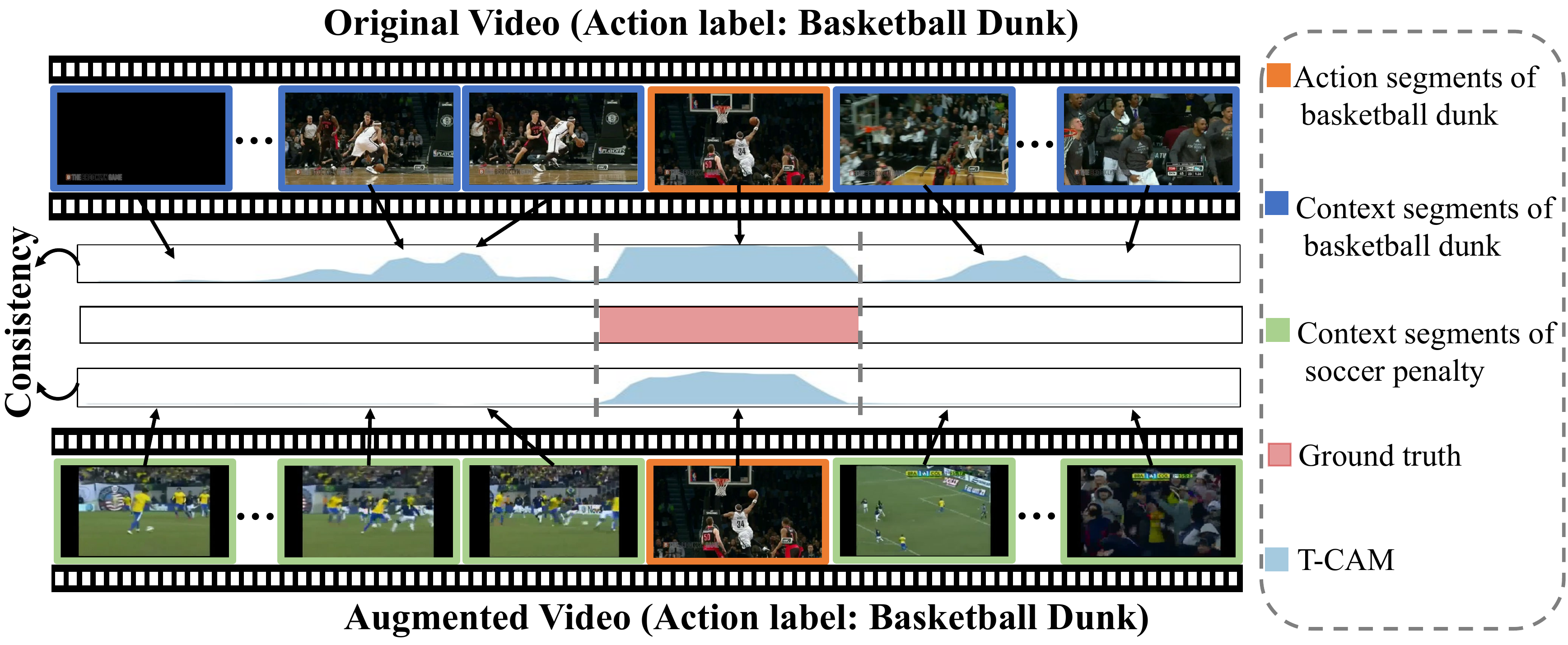}
\end{center}
   \caption{{Schematic of the semantic consistency constraint. Action instances tend to be accompanied by some inherent temporal context, \emph{i.e.} co-scene actions with the similar scenes as positive actions, such as the action `basketball dunk' and its context `dribble on the basketball court'. Similar scenes and {a} high correlation of co-scene actions and positive actions {lead} to confusion in the network, causing false positive activations in T-CAM. Our proposed temporal context augmentations and consistency constraint aim to randomly transform the temporal context of an action, so as to break the correlation between positive actions and co-scene actions to force the model to focus more on the action itself.}} 
\label{fig:mot}
\vspace{-0.5cm}
\end{figure}

Current WTAL methods such as \cite{2020WeaklyEM,yang2021uncertainty} consider {generating} frame-level pseudo-labels and turn the problem into full supervision to separate the action and its temporal contexts in the video. 
{ \cite{nguyen2019weakly} and \cite{lee2019background} generate background suppressed T-CAM according to a background suppress attention and force the context segments to belong to the background class after background suppression to help separate the action segments and background segments in the video. }
In spite of the promising performance, due to the co-scene actions in temporal contexts are similar and closely related to positive actions, it is still difficult to distinguishing positive actions from their inherent co-scene actions. In addition, since the classification model only focuses on the most distinguishable regions, existing methods still suffer from incomplete action localization.
Specifically, with only classification loss, the model would tend not only to focus on action itself but also action-related scenes. Hence those other actions in the action-related scenes (\emph{i.e.}, the scene same as positive actions) may be used to distinguish class label and be misclassified as positive actions by the model.
For example, the upper part of Figure \ref{fig:mot} shows a video that contains an action of `Basketball Dunk' and its corresponding T-CAM generated by a pre-trained baseline WTAL model. 
The action `Basketball Dunk' always occurs on the basketball court in the datasets, and is accompanied by some inherent co-scene actions, \emph{i.e.} those other actions that also occur on the basketball court such as dribbling or other sports that occur in the same scene, \emph{etc}. 
Different from static background segments, these co-scene actions are closely related to positive actions and have similar scenes. 
Hence the trained model may misidentify those co-scene actions, causing the model to yield confounding bias. As shown in the upper part of Figure \ref{fig:mot}, the model may also get relatively high prediction scores on the co-scene actions related to positive action. 

Different from the above methods, in this paper, we explore the self-supervised signal for WTAL through proposing an inter-video temporal context augmentation strategy ($Inter$-TCA), which could break the correlation between positive actions and their co-scene actions. 
As shown in the bottom half of Figure \ref{fig:mot}, the temporal context of action `Basketball Dunk' is replaced by the temporal context segments of another randomly selected `Soccer Penalty' video to generate an augmented video according to a pseudo-label generated by a pre-trained WTAL model.
The new generated augmented video still retains the action segment in the original video, \emph{i.e.} the `Basketball Dunk', but the correlation between the action and its co-scene actions has been broken.
Therefore, only the positive action segments in {the} augmented video can help to distinguish {different} class labels.
Then, a semantic consistency constraint is proposed to minimize the discrepancy between the background suppressed T-CAM of the original and augmented video, to suppress the co-scenes actions and force the model {to pay} more attention to the positive actions.
{Besides, we also design another intra-video temporal context augmentation ($Intra$-TCA) to obtain within class variations of the videos, which is implemented by changing the location of action instances within the video to change the temporal context of actions.}

Moreover, the temporal context augmentations destroy the correlation between the original action and {the} temporal context. 
We find that simply applying consistency constraint would cause the model to over-focus on the most discriminative parts and affect the completeness of localization. 
Hence, we boost the semantic consistency constraint in a bidirectional manner to suppress co-scene actions while maintaining the stability and integrity of positive actions prediction.  
Specifically, the semantic consistency constraint would be designed as a cross-supervision between the predicted background suppressed T-CAM of the original video and the augmented video. 
Additionally, a comprehensive T-CAM by aggregating multiple transformations of the same video's prediction is used in the bidirectional semantic consistency constraint (Bi-SCC) to further improves the integrity of positive actions prediction.  
Finally, to improve the quality of pseudo-labels produced by the pre-trained base model, we design an iterative process to update the parameters of the base model. Furthermore, the proposed Bi-SCC can be directly applied to the current methods and improve the performance of them.

The contributions are summarized as follows:
\begin{enumerate}
\item {We design two temporal context augmentation strategies and a semantic consistency constraint, which can break the correlation between positive action instances and their co-scenes actions, driving the model to focus more on action instances.}

\item {We boost the designed semantic consistency constraint in a bidirectional mutual supervision manner {to improve} the completeness of the action localization and the proposed method can be easily generalized to other current methods.}

\item {Extensive experiments show the effectiveness of our method on two public datasets, and the results show that our method can improve the performance of state-of-the-art approaches.}
\end{enumerate}    

\section{Related Work}
\label{sec:related work}

\subsection{{Action Recognition}}
\label{sec:AR}
{Action recognition aims to identify action categories in trimmed videos and is the basis for many video processing tasks.} 
{Current deep-learning-based methods could be roughly categorized into two types.}
{Two-steam networks \cite{simonyan2014two,feichtenhofer2016convolutional} utilize two separate CNNs to separately model appearance and motion information in videos with a late fusion.}
{3D CNN stack 3D convolutions to jointly model temporal and spatial semantics, such as C3D \cite{tran2015learning} and I3D \cite{DBLP:conf/cvpr/CarreiraZ17}.}
{FSNet \cite{zheng2021collaborative} proposes to boost the representative ability of spatiotemporal descriptors via spatiotemporal pyramid features.
Zheng et al. \cite{zheng2020global} propose to make full use of both significant local details and global information to boost action recognition.
Besides, SSTSA \cite{alfasly2022effective} proposes an effective Transformer-based network via recent strong vision transformers to jointly encode the spatial and temporal features.}
{
In addition to the above RGB-based methods, there are some skeleton-based methods \cite{li2021memory,zhang2019graph}.
In this paper, we use the action recognition model as the feature extraction module for the weakly supervised temporal action localization task.
}
%
\subsection{Weakly Supervised Temporal Action Localization}
\label{sec:w-tal}


\begin{figure*}[!t]
\setlength{\abovecaptionskip}{0.cm}
\setlength{\belowcaptionskip}{-0.cm}
\begin{center}
    \includegraphics[width=1\linewidth]{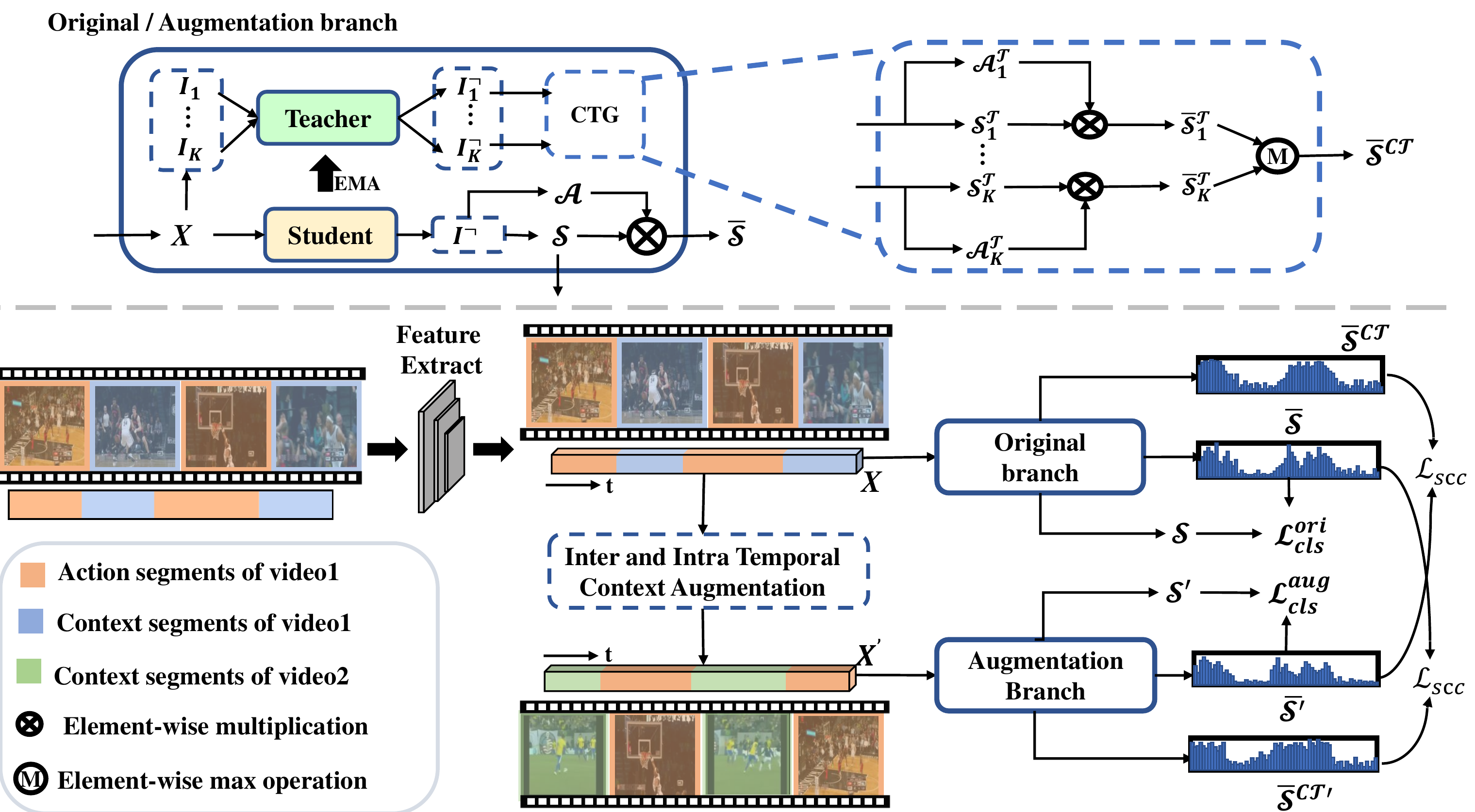}
\end{center}
   \caption{{Illustration of the proposed framework. In this figure, {the Temporal Context Augmentation  contains two strategies we designed,} $Inter$-TCA and $Intra$-TCA. ${I}$ indicates the proposed $Intra$-TCA, and ${I}_1,...,{I}_K$ means $K$ different $Intra$-TCA. ${I}^{\lnot}$ is the inverse operation of the $Intra$-TCA, which aims to reset the shuffled action instance. The original branch input is the original video feature, while the augmentation branch takes the temporal context augmented video feature as input. Furthermore, a comprehensive T-CAM generation (CTG) module is used to generate comprehensive background suppressed  T-CAM. The semantic consistency constraint {is} imposed between the two branches.}}
\label{fig:fra}
\vspace{-0.5cm}
\end{figure*}

{Weakly supervised temporal action localization requires video-level labels only. Due to the lack of precise boundary labels, most advanced WTAL methods fall into a multiple-instance-learning (MIL) framework to tackle the WTAL task. 
For example, UntrimmedNet   \cite{wang2017untrimmednets} proposes a selection module to rank clip proposals and locate action instances.
{AdapNet \cite{zhang2020adapnet} proposes to fully leverage the trimmed videos as external resources to further improve the analytical performance.}
{\cite{chen2019relation} proposes relation attention to exploit informative relations among action proposals based on their appearance or optical flow features. Breaking Winner-Takes-All \cite{zeng2019breaking} focus on carefully designing adversarial erase strategy to model action completeness.}
W-TALC \cite{paul2018w}, 3C-Net \cite{narayan20193c} add the metric learning loss to the MIL framework to decrease intra-class variations and increase inter-class difference.
{\cite{Liu_2021_CVPR} proposes a temporal smoothing PCA-based deconfounder to exploit the unlabelled background to model an observed substitute for the unobserved confounder, to remove the confounding effect. \cite{li2022exploring} 
using contrastive learning to distinguish foreground actions and backgrounds.}
{In BM} \cite{nguyen2019weakly} and BaS-Net \cite{lee2019background}, an additional background class and a background suppression loss {are} applied to help separate the action segments and background segments in the video. 
Meanwhile, \cite{yang2021uncertainty} consider obtaining frame-level pseudo-labels to improve the quality of T-CAM. {In addition to the above methods,\cite{yang2021background} propose to use single-frame annotations for weakly supervised action localization and \cite{Gong_2020_CVPR} proposes unsupervised action localization by clustering unlabeled videos.}
Besides, Su \emph{et al.} \cite{su2021improving}, and Hong \emph{et al.} \cite{hong2021cross} impose consistency constraint between videos of different temporal resolutions or between RGB and Flow. 
In spite of the promising performance, these methods are limited by the context confusion problem.
In this paper, we design a temporal context augmentation strategy to break the correlation between actions and their co-scene actions in the temporal context.
Additionally, we propose a bidirectional semantic consistency constraint to improve the performance of the WTAL model. }

\subsection{Self-Supervised Learning}
\label{sec:SSL}
{
Self-supervised learning leverages unlabeled data to make the model learn intrinsic knowledge from data. Currently, many self-supervised learning methods have been proposed to learn better feature representations. 
It is often useful to exploit the self-supervised learning strategy for learning better representation when we lack full supervision \cite{zhang2020weakly,zhang2021weakly}. For example, in weakly supervised semantic segmentation, spatial multi-resolution information has been used for equivariant consistency constraint \cite{wang2020self}. Gong \emph{et al.} \cite{gongself} proposed self-supervised equivariant transform consistency constraint. Besides, Su \emph{et.al} \cite{su2021improving} utilizes temporal multi-resolution information from the weakly-supervised temporal action localization task for pseudo label generation and representation learning. However, for WTAL, adjusting the video as a whole and maintaining the same contextual background segments will not significantly help the model to mine more accurate action areas. In contrast, our work takes advantage of the semantic consistency constraints to suppress the co-scene {action to improve} the quality of T-CAM to improve the performance of action localization. 
}

\section{Methodology}
\label{sec:methodology}
In this section, we first introduce the problem definition of WTAL in Section \ref{subsec: PD}. {Later we introduce the video feature extraction method in Section \ref{FE}. Finally we introduce the main pipeline of the proposed framework in Section \ref{subsec:MP}.}

\subsection{Problem Definition}
\label{subsec: PD}
In the WTAL task, we are provided a set of $N$ untrimmed videos defined as $\{{V_n}\}_{n=1}^N$, and all of them are annotated with their corresponding video-level labels $\{{y_n}\}_{n=1}^N$. The label $y_n$ of the $n$-th video is a binary vector indicating the presence/absence of each action.  Each video $V$ is a collection of segments: ${V} = \{ {v_t}\}_{t=1}^{T}$, and $T$ is the number of segments in the video. During the test, we want to be able to predict a sequence of actions $\{c_j, s_j, e_j, {conf_j}\}$ in an input video. Each action is represented by the action start time $s_j$, end time $e_j$, action category $c_j$, and confidence score {$conf_j$.}

\subsection{Video Feature Extraction}
\label{FE}
Following recent methods \cite{paul2018w, lee2019background, islam2021hybrid, huang2021foreground, li2021multi, ding2021kfc, Ding_2021_ICCV}, the segment-level RGB feature sequence $\mathbf{X}^r\in R^{T \times D}$ and optical flow feature sequence $\mathbf{X}^o\in R^{T\times D}$ are extracted by the I3D network \cite{DBLP:conf/cvpr/CarreiraZ17} pre-trained on the Kinetics dataset \cite{article2017} without any fine-turning. $D$ represents the dimension of the feature. Apart from this, since different videos vary in temporal length, We interpolate the extracted segment-level features so that all videos contain the same number of segments. Finally, the extracted features $\mathbf{X} = [\mathbf{X}^r;\mathbf{X}^o] \in R^{T \times 2D}$ will be used in the next module. 

\subsection{Main Pipeline}
\label{subsec:MP}

\begin{figure}[!t]
\begin{center}
    \includegraphics[width=1\linewidth]{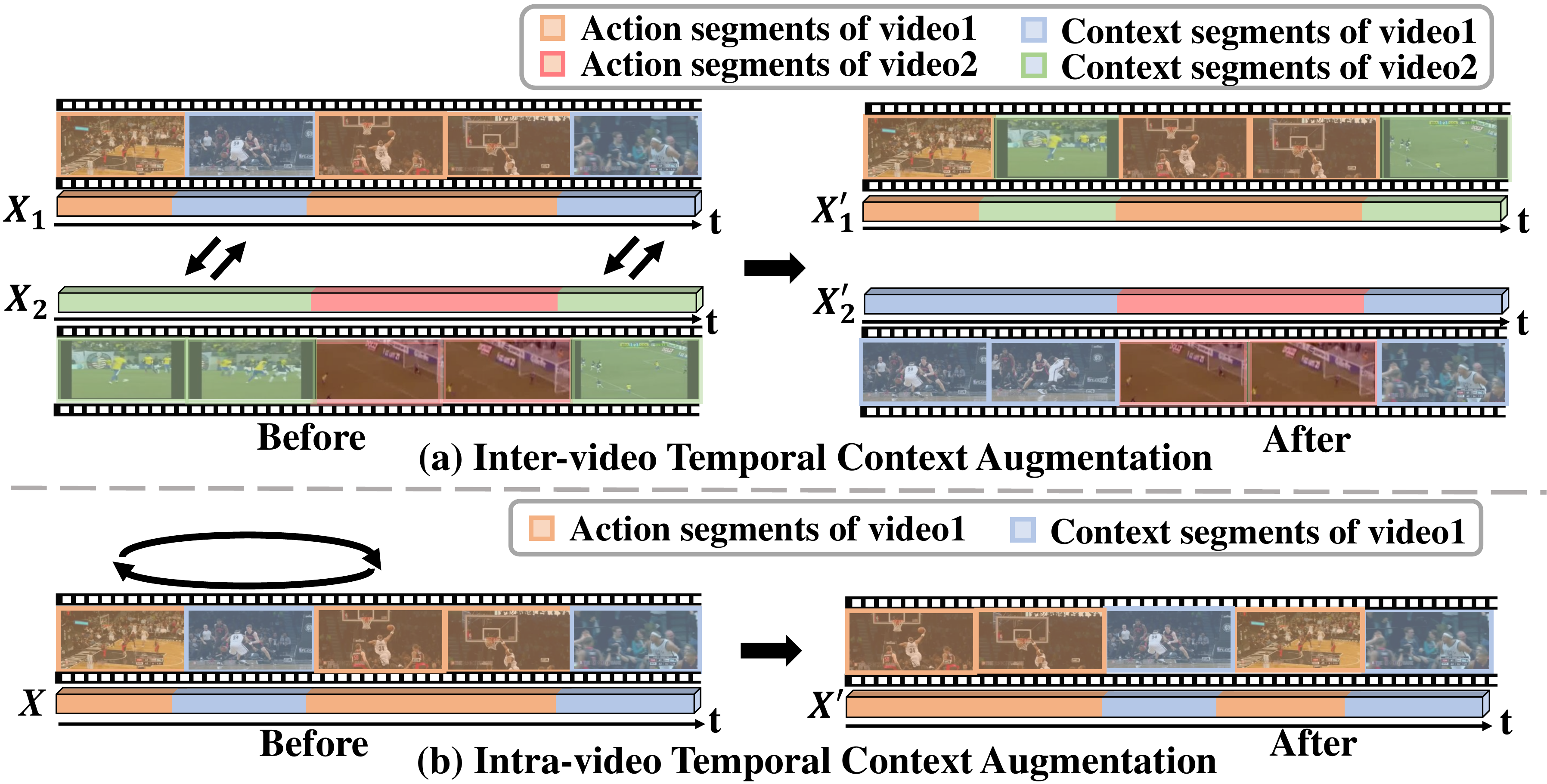}
\end{center}
   \caption{{Illustration of the two proposed temporal context augmentation strategies.{(a) shows the proposed Inter-video Temporal Context Augmentation ($Inter$-TCA) strategy. The $Inter$-TCA augmentation randomly combines the collected pseudo positive action segments with pseudo temporal context segments, which would break the correlation between the actions of their co-scene action in the temporal context.} (b) shows the proposed  Intra-video Temporal Context Augmentation ($Intra$-TCA) strategy. The $Intra$-TCA strategy changes the temporal context of the action instance by changing the position of the action instance in the original video.}}
\label{fig:tca}
\vspace{-0.5cm}
\end{figure}
In this section, we elaborate on the proposed method. The illustration of the method is shown in Figure \ref{fig:fra}. {Following the recent WTAL methods \cite{islam2021hybrid,hong2021cross}, high-level feature representations of the RGB and optical flow video are extracted by the pre-trained I3D \cite{DBLP:conf/cvpr/CarreiraZ17} network at first. {Then the proposed inter-video context augmentation is applied to the input video features to break the correlation between action instances and their co-scene actions in the temporal context.} Later, the original video feature and the augmented video feature are input into two branches with the same structure. We adopt the traditional mean-teacher \cite{tarvainen2017mean} framework to build each branch, where the teacher network is updated through EMA from the student. Compared to the student network, the teacher network of each branch receives multiple video features that augmented by different intra-video temporal context augmentations {to further break the correlation between action and its temporal context}, and obtain comprehensive background suppressed T-CAM through a comprehensive T-CAM generation module (CTG). Then, we impose a semantic consistency constraint between the student output of the original branch and the teacher output of the augmentation branch to force the model to pay more attention to the action. Besides, we also {impose} the semantic consistency constraint between the teacher output of the original branch and the student output of the augmentation branch to improve the semantic consistency in a bidirectional supervised manner to ensure the integrity of positive actions.}

\subsubsection{Temporal Context Augmentation}
\label{subsec:TCA }

\paragraph{Action Instance Collection}
\label{subsec:AIC }
For temporal context augmentation of action segments in videos, in this section, we collect preliminary action instances in videos by a pre-trained baseline model at first. In specific, we first train a baseline network to generate a segment-level pseudo-label which indicates the probability that a segment belongs to any action. {The baseline model does not include the temporal context augmentation and semantic consistency constraints and the baseline model is still trained under weak supervision.} { The weakly-supervised training procedure for the baseline model is shown in Algorithm \ref{A:0}}

Then the probability that the segment belongs to anyone action above the threshold $\gamma$ should be considered as positive action segments. After the action instance collection process, we can obtain a segment-level pseudo-label, called the action instance mask, where the value of the pseudo positive action segment is 1 and the pseudo temporal context segment is 0. With the action instance mask, we obtain pseudo action instances and their temporal positions in the video.

\paragraph{Inter-video Temporal Context Augmentation}
\label{subsec:TCT}
Before training the network for the next step, we first perform temporal context augmentation by transforming the temporal context of action instances in the video. As shown in Figure \ref{fig:tca} (a), we show the result of the Inter-video temporal context augmentation ($Inter$-TCA) of several action videos. To be specific, we select two videos from the dataset randomly and extract pseudo action and pseudo temporal context segments of the two videos respectively through the action instance mask generated in \ref{subsec:AIC } at first. 

After that, we swap the pseudo temporal context segments of the two videos to generate temporal context-augmented videos. Note that, extracting features from scratch is not necessary for the generated augmented videos. The features $\mathbf{X}^{'}$ of temporal context-augmented videos can be conveniently obtained by swapping features of temporal context segments of two original videos with the action instance mask.

Formally, for two randomly selected videos $V_1$ and $V_2$, their feature sequences are extracted as $\mathbf{X}_1,\mathbf{X}_2 \in R^{T\times 2D}$ according to section \ref{FE}. Then we apply action instance mask ${m}_1,{m}_2 \in R^{T\times1}$ to extract features of pseudo positive action segments $\mathbf{X}^a_1,\mathbf{X}^a_2$ and pseudo temporal context segments $\mathbf{X}^b_1,\mathbf{X}^b_2$:
\begin{equation} 
\begin{aligned}
\label{E1} 
\mathbf{X}^a_1 = {m}_1 * \mathbf{X}_1, & \mathbf{X}^a_2 = {m}_2 * \mathbf{X}_2,\\
\mathbf{X}^b_1 = (1-{m}_1) * \mathbf{X}_1,& 
\mathbf{X}^b_2 = (1-{m}_2) * \mathbf{X}_2,
\end{aligned}
\end{equation}
where $\mathbf{X}^a_1,\mathbf{X}^a_2,\mathbf{X}^b_1,\mathbf{X}^b_2 \in R^{T\times 2D}$. Assuming that the number of context segments contained in videos $V_1$ and $V_2$ is $N^b_1$ and $N^b_2$ respectively, we collect all temporal context segments from $\mathbf{X}^b_1$, and $\mathbf{X}^b_2$ to generate $\hat{\mathbf{X}}^b_1 \in R^{N^b_1 \times 2D}$, and $\hat{\mathbf{X}}^b_2 \in R^{N^b_2 \times 2D}$. Finally, we combine the pseudo positive action segments of video $v_1$, $v_2$ with the pseudo temporal context segments of video $v_2$, $v_1$ to obtain the temporal context-augmented videos, and the feature of the new video $\mathbf{X}^{'}_1$ and $\mathbf{X}^{'}_2$ can be obtained by:
\begin{equation} 
\begin{aligned}
\label{E2} 
\mathbf{X}^{'}_1 = \mathbf{X}^a_1 + (1-{m}_1) * U(\hat{\mathbf{X}}^b_2), \\
\mathbf{X}^{'}_2 = \mathbf{X}^a_2 + (1-{m}_2) * U(\hat{\mathbf{X}}^b_1), \\
\end{aligned}
\end{equation}
where, $U$ represents the upsampling operation, which upsamples $\hat{\mathbf{X}}^b$ from $N^b$ to $T$ in the time dimension. 

\paragraph{Intra-video Temporal Context Augmentation Strategy}
\label{subsec: IAS}
Intra-video Temporal Context Augmentation ($Intra$-TCA) strategy aims to perform temporal context augmentation of action instances within a video. As shown in Figure \ref{fig:tca} (b), we show the result of $Intra$-TCA of several videos. To be specific, the $Intra$-TCA strategy collects the action instances in the video according to the acquired action instance mask and then changes the temporal context of the action instance by changing the location of the action instance in the video. Note that the $Intra$-TCA strategy can also be conveniently implemented at the feature level.

Formally, we apply the action instance mask to collect features of action instances $\{\mathbf{x}^a_{[s_i,e_i]}\}_{i=1}^M$ from a video feature $\mathbf{X}=\{\mathbf{x}_t\}_{t=1}^T$, where  $M$ is the numeber of pseduo-action instances collected in a video. For the $Intra$-TCA strategy, we randomly select two pseudo-action instance $\mathbf{x}^a_{[s_1,e_1]}$ and $\mathbf{x}^a_{[s_2,e_2]}$ in a video, inflate their boundaries slightly and swap their positions in the video. For video with only one action instance, we change its location in the video directly.

\subsubsection{T-CAM Generation}
\label{subsec:T-CAM Generation}
Temporal attention weights indicate the probability that each segment belongs to any action, and those segments with low attention values are more likely to belong to the background segment. Here, following previous works \cite{lee2019background,islam2021hybrid,nguyen2018weakly}, an attention unit is also used to generate class-agnostic attention:
\begin{equation} 
\begin{aligned}
\label{E4} 
\mathbf{A} = Att(\mathbf{X})
\end{aligned}
\end{equation}
where $Att$ is the attention unit with three convolution layers.

After obtaining the video features and class-agnostic attention weights, T-CAM $\mathbf{S} \in R^{T\times (C+1)}$ and background suppression T-CAM $\mathbf{\bar{S}} \in R^{T\times (C+1)}$ can be obtained by a classifier $f$ that contains a non-local \cite{wang2018non} layer and four convolution layers, where the $(C+1)$-th class is the background class.

\begin{equation} 
\begin{aligned}
\label{E5} 
\mathbf{S} = f(\mathbf{X}), \quad \mathbf{\bar{S}} = \mathbf{A} * \mathbf{S}.
\end{aligned}
\end{equation}

 
\subsubsection{semantic consistency constraint}
\label{sec:SCC}
Since the temporal context of the augmented video is not closely related to the action, it will be predicted as an extra background class with a high probability in the background-suppressed T-CAM without being misidentified as an action. Thus the background-suppressed T-CAM of the augmented video could be used to supervise that of the original video with semantic consistency constraints, forcing the model to predict the context of the original video containing the co-scene action as addition background classes rather than actions. Besides, the temporal context augmentation strategy does not change the internal segments' order of the original action instance, so the background-suppressed T-CAM before and after the $Inter$-TCA strategy should be consistent.

Formally, given a feature $\mathbf{X}$ of an untrimmed video and the temporal context augmentation strategy $Inter$-TCA in \ref{subsec:TCT}, the new video feature $\mathbf{X}^{'}$ will be generated by replacing the original temporal context. Besides, $Intra$-TCA can be also used to augment the temporal context within the video. Similiar as \cite{cheng2022hybrid,li2021adaptively,zhang2019deep}, we adopt two neural network to form a mean-teacher network as a baseline, where both teacher and student networks are T-CAM generation modules in \ref{subsec:T-CAM Generation}. The original video feature $\mathbf{X}$ is fed into the student network to get the background-suppressed T-CAM $\mathbf{\bar{S}}$ while the background-suppressed T-CAM $\mathbf{\bar{S}}^{\mathcal{T}'}$ of augmented video feature $\mathbf{X}^{'}$ can be generated from the teacher network. Note that the superscript $\mathcal{T}$ here represents the output from the teacher network. In this paper, the semantic consistency constraint is formulated by using the KL divergence:

\begin{equation} 
\begin{aligned}
\label{E7} 
 L_{SCC} = KL(\mathbf{\bar{S}}^{\mathcal{T}'} || \mathbf{\bar{S}}).
\end{aligned}
\end{equation}

The temporal context augmentation can break the dependency between action instances and contextual background. The semantic consistency loss can provide an additional constraint between the original and temporal context augmentation T-CAMs. In this way the model can be made to pay more attention to the internal information of action instances to predict the class they belong to, thereby improving the quality of the generated T-CAM.

\begin{table*}[h]
    \setlength{\abovecaptionskip}{-0.1cm}
    \caption{Action localization results in the THUMOS14 dataset. This table reports the mAP values in IoU@(0.1-0.7). ``Weak+'' means that the method uses extra supervision. Compared with the previous weak supervision method, our method achieves better performance even close to the full supervision method.}
    \begin{center}
    \label{tab:1}
    \begin{tabular}{c|c|m{0.75cm}<{\centering}m{0.75cm}<{\centering}m{0.75cm}<{\centering}m{0.75cm}<{\centering}m{0.75cm}<{\centering}m{0.75cm}<{\centering}m{0.75cm}<{\centering}|m{1cm}<{\centering}m{1cm}<{\centering}}
    \hline\hline
    
        \multirow{2}{*}{Supervision} &
         \multirow{2}{*}{Model} &
        \multicolumn{7}{c|}{mAP@IoU(\%)} &
        \multicolumn{2}{c}{AVG}  \\  \cline{3-11}
         ~ & ~ &0.1 &0.2 & 0.3 & 0.4 & 0.5 & 0.6 & 0.7 & 0.1:0.5 & 0.1:0.7
         \\ \hline \hline
        Full & AGCN \cite{li2020graph} AAAI2020  &59.3& 59.6 & 57.1 &51.6& 38.6 & 28.9 & 17.0 & 53.2 & 44.6\\ 
        Full & GTAN \cite{long2019gaussian} CVPR2019 &69.1 & 63.7& 57.8 & 47.2 & 38.8 & - &-& 55.3 &-\\ 
        Full & P-GCN \cite{zeng2019graph}  ICCV2019  & 69.5& 67.8 & 63.6 & 57.8 & 49.1 &  -& - & 61.6&- \\\hline
        Weak+ & SF-Net \cite{Ma2020SF} ECCV2020  &68.3 & 62.3& 53.2 & 40.7 & 29.3 & 18.4 & 9.6&51.2&41.2 \\
        Weak+ & Ding \emph{et al.} \cite{2020Weakly}  axriv2020  &61.6 & 55.8 & 48.2 &39.7 & 31.6 & 22.0 & 13.8&  47.4 & 39.0 \\ 
        Weak+ & Ju \emph{et al.} \cite{ju2021divide} CVPR2021  & 72.8 & 64.9 & 58.1 & 46.4 & 34.5 & 21.8 & 11.9 & 55.3 & 44.3 \\\hline
        Weak & BaS-Net \cite{lee2019background} AAAI2020 &58.2&52.3& 44.6 & 36.0 & 27.0 & 18.6 & 10.4&43.6&35.3 \\ 
        Weak & EM-MIL \cite{2020WeaklyEM} ECCV2020  &59.1&52.7& 45.5 & 36.8 & 30.5 & 22.7 & \textbf{16.4} & 45.0 & 37.7\\ 
        Weak & AAAI2021  HAM-Net \cite{islam2021hybrid} & 65.4 & 59.0 & 50.3 & 41.1 & 31.0 & 20.7 & 11.4 & 46.4& 39.8 \\
        Weak & Gong \emph{et al.} \cite{gongself} IJCAI2021 & 64.8 & 58.4 & 50.8 & 42.2 & 32.9 & 21.0 & 10.1 & 54.0 & 43.0\\
        Weak & UGCT \cite{yang2021uncertainty} CVPR2021  &69.2 & 62.9 & 55.5 & 46.5& 35.9 & 23.8& 11.4 & 54.0 & 43.6  \\
        Weak & FACNet \cite{huang2021foreground}  ICCV2021  & 67.6 & 62.1 & 52.6 & 44.3 & 33.4 & 22.5 & 12.7 & 52.0 & 42.2  \\
        Weak &CO2-Net \cite{hong2021cross} ACM MM2021  & 70.1 & 63.6 & 54.5&45.7 & 38.3 & 26.4 & 13.4 & 54.4 & 44.6  \\
        Weak & ACG-Net \cite{yang2021acgnet} AAAI2022   &68.1&  62.6 &53.1&44.6 &34.7&22.6&12.0 &52.5 &42.4 \\
        Weak & FTCL \cite{gao2022fine} CVPR2022 & 69.6 & 63.4 & 55.2 & 45.2 & 35.6 & 23.7 & 12.2 & 53.8 & 43.6\\
        Weak & ASM-Loc \cite{he2022asm} CVPR2022 & 71.2 & 65.5 & 57.1 & 46.8 & 36.6 & 25.2 & 13.4 & 55.4& 45.1 \\
        Weak &Huang \emph{et al.}   \cite{huang2022weakly} CVPR2022 & 71.3 & 65.3& 55.8 & 47.5 & 38.2& 25.4 & 12.5 & 55.6 & 45.1 \\
        Weak & Ours & \textbf{71.7} & \textbf{66.9}& \textbf{57.2} & \textbf{48.0} & \textbf{40.4} & \textbf{27.5} & 14.4 & \textbf{56.9}&\textbf{46.6} \\
        \hline\hline
    \end{tabular}
    \end{center}
\vspace{-0.5cm}    
\end{table*}

\begin{algorithm}[t]

\caption{{The Weakly-Supervised Training Procedure for the Baseline Model}}
\label{A:0}
\hspace*{0.02in} {\bf Input:}
  {Video features of training set: ${X}_{i=1}^N$}, 
  {Video labels of training set: $y_{i=1}^N$ },
  {The number of iterations of the algorithm: $Itr$}
\begin{algorithmic}[1]
\For{$itr=1,...,Itr$}
    \State {Input $X$ to the student network composed of T-CAM generation module (Section \ref{subsec:T-CAM Generation}) to generate T-CAM $S$ and background suppressed T-CAM $\bar{S}$ according to Eq. \ref{E4} and Eq. \ref{E5}.}
    \State {Input $X$ to the teacher network composed of T-CAM generation module (Section \ref{subsec:T-CAM Generation}) to generate another background suppressed T-CAM $\bar{S}^{\mathcal{T}}$ according to Eq. \ref{E4} and Eq. \ref{E5}.}
    \State {Optimize the student network by Eq. \ref{E10}-Eq. \ref{E14} (Section \ref{sec:LF}) and Eq. \ref{E7} (Section \ref{sec:SCC}).}
    \State {Update the teacher network’s parameters from student network’s parameters by the exponential moving average (EMA) strategy.}
\EndFor
\end{algorithmic}

\end{algorithm}

\subsection{Bidirectional semantic consistency constraint}
\label{subsec:biscc}

To break the correlation between the positive actions and their co-scene actions, the temporal context augmentations destroy the dependency between the original actions and their temporal context. We find that simply applying consistency would cause the model to over-focus on the most discriminative parts of the video and affect the completeness of localization. Hence we boost the semantic consistency constraint in a bidirectional manner to suppress co-scene actions while maintaining the stability and integrity of positive actions prediction. Furthermore, considering that different transformations of the same action video will produce different predictions, combining multiple transforms of the same video to generate a comprehensive T-CAM always covers a more comprehensive part of actions than a single one. Hence we also utilize the comprehensive background suppressed T-CAM to ensuring the integrity of positive actions.
Specifically, in this paper, {we adopt the traditional teacher-student network \cite{ tarvainen2017mean} to compose our model.
The model proposed in this paper contains an original branch and an augmentation branch, both of which contain a teacher network and a student network.}
{Specifically, in this paper, the original branch includes a teacher-student network. The student network of the original branch  takes an original video feature $\mathbf{X}$ as input while the inputs of the teacher network of the original branch are multiple video features transformed by $Intra$-TCA in \ref{subsec: IAS}. In the teacher network of the original branch,} $K$ different inputs including $\mathcal{I}_1(\mathbf{X}),...,\mathcal{I}_K(\mathbf{X})$ are fed into the same modules to get multiple background-suppressed T-CAM $\mathbf{\bar{S}}^{\mathcal{T}}_1,...,\mathbf{\bar{S}}^{\mathcal{T}}_K$, where $\mathcal{I}()$ means the proposed intra-video temporal context augmentation strategy. Then the comprehensive T-CAM {generation} module would generate comprehensive background suppressed T-CAM,

\begin{equation} 
\begin{aligned}
\label{E8} 
{
\mathbf{\bar{S}}^{C\mathcal{T}} = \mathcal{G}({\mathcal{I}_{1}^{\lnot}}(\mathbf{\bar{S}}^\mathcal{T}_1),...,{\mathcal{I}_{K}^{\lnot}}(\mathbf{\bar{S}}^\mathcal{T}_k)),
}
\end{aligned}
\end{equation}
where $\mathcal{G}$ is the comprehensive T-CAM generation module, which computes the comprehensive background suppressed T-CAM by element-wise max operator. Besides, there is an inverse operation of the $Intra$-TCA {$\mathcal{I}^{\lnot}()$} in comprehensive T-CAM generation module, which restore the shuffled action instance in the $Intra$-TCA to its original position in the video sequence. 
{On the other hand, the augmentation branch also contains a teacher-student network. The student network of the augmentation branch takes the video features after $Inter$-TCA $X^{'}$ as input, while the teacher of the augmentation branch network takes multiple video features after $Inter$-TCA and $Intra$-TCA as input, ~\emph{i.e.} {$\mathcal{I}_1(\mathbf{X}^{'}),...,\mathcal{I}_K(\mathbf{X}^{'})$.}} Similarly, in temporal context augmentation branch, the comprehensive background-suppressed T-CAM $\mathbf{\bar{S}}^{C\mathcal{T}'}$ can be also obtained by averaging $\mathbf{\bar{S}}^{\mathcal{T}'}_1,...,\mathbf{\bar{S}}^{\mathcal{T}'}_K$ {generated by $\mathcal{I}_1(\mathbf{X}^{'}),...,\mathcal{I}_K(\mathbf{X}^{'})$. }

Finally, we boost the semantic consistency constraints in a bidirectional manner by cross supervising the original and augmented video. The Eq. \ref{E7} can be improved into the following formula:
\begin{equation} 
\begin{aligned}
\label{E9} 
 L_{Bi-SCC} = KL(\mathbf{\bar{S}}^{C\mathcal{T}} || \mathbf{\bar{S}}^{'}) + KL(\mathbf{\bar{S}}^{C\mathcal{T}'} || \mathbf{\bar{S}}).
\end{aligned}
\end{equation}

\subsection{Loss Formulation} 
\label{sec:LF}
In addition to the semantic consistency loss, in this work, we apply the widely used top-k multiple-instance learning loss \cite{lee2019background} on both T-CAM $\mathbf{S}$ and background-suppressed T-CAM $\mathbf{\bar{S}}$ to train student network of the original branch. In the specific, the temporal pooling is performed by aggregating the top-k values from the temporal dimension for each class. For class label $j$: 

\begin{equation}
\label{E10} 
v_{j}=\max _{\substack{l \subset\{1, \ldots, T\} \\|l|=k}} \frac{1}{k} \sum_{i \in l} \mathbf{S}_{i}(j), \quad
 v_{j}^{supp}=\max _{\substack{l \subset\{1, \ldots, T\} \\|l|=k}} \frac{1}{k} \sum_{i \in l} \mathbf{\bar{S}}_{i}(j), 
\end{equation}
where $v_{j}$ and $v_{j}^{supp}$ are video-level class score generated by original T-CAM and background-suppressed T-CAM respectively. Then, the video-level classification probability $p_{j}$ and $p_{j}^{supp}$ can be obtained by:
{
\begin{equation}
\label{E12} 
p_{j}=\frac{\exp \left(v_{j}\right)}{\sum_{i=1}^{C+1} \exp \left(v_{i}\right)}, \quad 
p_{j}^{supp}=\frac{\exp \left(v_{j}^{supp}\right)}{\sum_{i=1}^{C+1} \exp \left(v_{i}^{supp}\right)}.
\end{equation}
}
Finally, we calculate the final loss function $L_{cls}^{ori}$:
{
\begin{equation}
\label{E13}
\mathcal{L}_{cls}^{ori}=-(\sum_{j=1}^{C+1} y_{j} \log \left(p_{j}\right)+\sum_{j=1}^{C+1} y_{j}^{supp} \log \left(p_{j}^{supp}\right)),
\end{equation}
where the additional C+1-th background class is 0 in $y_{j}^{supp}$ and 1 in $y_{j}$. 
}
Similiary, to train student network of the augmentation branch, the top-k multiple-instance learning loss also apply to $\mathbf{S}^{'}$ and $\mathbf{\bar{S}^{'}}$, denoted as $L_{cls}^{aug}$ according to Eq. \ref{E10}-\ref{E13}.
Besides, following \cite{islam2021hybrid,hong2021cross}, we also utilize normalization loss $L_{norm}$, guide loss $L_{guide}$ and co-activity similarity loss $L_{cas}$ to train the student network of both the original branch and the temporal context augmentation branch. The final loss function in this paper is 
\begin{equation} 
\begin{aligned}
\label{E14} 
L^{ori} = L_{cls}^{ori} + L_{norm}^{ori} + L_{guide}^{ori} + L_{cas}^{ori},
\end{aligned}
\end{equation}

\begin{equation} 
\begin{aligned}
\label{E15} 
L^{aug} = L_{cls}^{aug} + L_{norm}^{aug} +  L_{guide}^{aug} + L_{cas}^{aug}.
\end{aligned}
\end{equation}

The final objective function for whole framework is:
\begin{equation} 
\begin{aligned}
\label{E16} 
L = L^{ori} + L^{aug} + \alpha L_{Bi-SCC}.
\end{aligned}
\end{equation}
Here, $\alpha$ are hyper-parameters. By optimizing the final objective function, our framework can generate more accurate T-CAMs. Besides, we adopt an iterative scheme to update the parameters of the pre-trained base model. Finally, to improve the quality of the pseudo label in Section \ref{subsec:AIC }, we update the parameters of the basic pre-trained WTAL model by the student network of the original branch. The overall process of the proposed algorithm is shown in Algorithm \ref{A:1}

\begin{algorithm}[t]

\caption{The Overall Framework} 
\label{A:1}
\hspace*{0.02in} {\bf Input:}
  Video features of training set: ${X}_{i=1}^N$ \\
  Video labels of training set: $y_{i=1}^N$ \\
  A base pre-trained WTAL model: $\mathcal{F}$ \\
  The number of iterations of the algorithm: $Itr$
\begin{algorithmic}[1]
\For{$itr=1,...,Itr$}
    \State Input the $X$ to the $\mathcal{F}$ to get segment-level pseudo lable according to \ref{subsec:AIC }.
    \State Generate augmented video features $X^{'}$ according to \ref{subsec:TCT} and \ref{subsec: IAS}.
    \State Input $X$ to the student of the original branch to generate T-CAM $S$ and background suppressed T-CAM $\bar{S}$.
    \State Input $X$ to the teacher of the original branch and the CTG module to generate comprehensive background suppressed T-CAM $\bar{S}^{C\mathcal{T}}$ according to \ref{subsec:biscc}
    \State {Input $X^{'}$ to the student of the augmentation branch to generate T-CAM $S^{'}$ and background suppressed T-CAM $\bar{S}^{'}$.}
    \State Input $X^{'}$ to the teacher of the augmentation branch and the CTG module to generate comprehensive background suppressed T-CAM $\bar{S}^{C\mathcal{T}'}$ according to \ref{subsec:biscc}
    \State Impose bidirectional semantic consistency constraint between $\bar{S}^{C\mathcal{T}'}$ and $\bar{S}$, $\bar{S}^{C\mathcal{T}}$ and $\bar{S}^{'}$.
    \State Train the student network of both original and augmentation branch by the loss functions in \ref{sec:LF}.
    \State Update the parameters of $\mathcal{F}$ by the student of original network.
\EndFor
\end{algorithmic}

\end{algorithm}

\subsection{Temporal Action Localization} 
\label{sec:TAL}
During the test stage, only the student network of the original branch is used to generate the proposals. Specifically, we follow the process of \cite{islam2021hybrid,hong2021cross}. Firstly, we select those classes with video-level category scores above a threshold for generating proposals. Then for the selected action classes, we drop the background snippets and obtain the class-agnostic action proposals by thresholding the attention weights and selecting the continuous components of the remaining snippets. As we said in \ref{subsec: PD}, the obtained candidate action proposals can be denoted as $\{c_j, s_j, e_j, {conf_j}\}$ including action start time $s_j$, end time $e_j$, action category $c_j$, and confidence score $conf_j$. For the confidence score $conf_j$, we follow the AutoLoc \cite{shou2018autoloc} to calculate the outer-inner score of each action proposal. Note that for calculating class-specific scores, we use the background-suppressed T-CAM of $c$-th category $\mathbf{\bar{S}}_c$ as \cite{lee2019background,islam2021hybrid,hong2021cross},
\begin{equation}
\begin{aligned}
\label{E17}
q_j &= Avg({\mathbf{\bar{S}}_c[s_j:e_j,c_j]})-Avg({\mathbf{\bar{S}}_c[s_j-l_j:e_j+l_j,c_j]}),\\
q_j &= q_j+0.2*\hat p_{cj}, \\
l_j &= 0.25*(e_j-s_j),
\end{aligned}
\end{equation}
where $\hat p_{cj}$ is the video-level score for class $c$. Finally, multiple thresholds strategy will be used to enrich the proposal set and non-maximum suppression will be used to remove duplicate proposals. 

\section{Experimental Results and Analysis}
\label{sec:Experimental Results and Analysis}
To verify the effectiveness of the proposed method, we evaluate our method on two datasets, \emph{i.e.}, THUMOS14 dataset \cite{THUMOS14}, and ActivityNet1.3 dataset \cite{caba2015activitynet}.

\subsection{Experimental settings}
\label{subsubsec: Datasets}
\subsubsection{THUMOS14} The THUMOS14 \cite{THUMOS14} dataset contains 200 validation videos and 213 test videos belonging to 20 categories with the time annotation of actions. Each video on the THUMOS14 dataset contains several action instances and the duration of action varies greatly, from less than a second to several minutes, making them highly challenging. In the same setting as \cite{paul2018w,nguyen2018weakly,narayan20193c,huang2022multi}, we adopt 200 validation videos as a training dataset and 213 test videos for testing. 

\subsubsection{ActivityNet}  
Compared with THUMOS14, ActivityNet \cite{caba2015activitynet} dataset offers a larger benchmark for temporal action localization with two versions. There are 10,024 training videos, 4,926 validation videos, and 5,044 testing videos with 200 action categories in the ActivityNet1.3. Following the setting of \cite{huang2022multi,huang2021foreground,yang2021uncertainty,huang2022weakly}, we adopt the training videos to train our model and verify our framework's performance in testing videos.

\begin{table}[!tbp]
    \centering
    \setlength{\abovecaptionskip}{-0.1cm}
    \caption{Action localization results in the ActivityNet1.3 dataset. The Avg column indicates the average mAP at IoU thresholds 0.5:0.05:0.95.}
    \begin{center}
    \begin{tabular}{m{3cm}<{\centering}|m{0.5cm}<{\centering}m{0.5cm}<{\centering}m{0.5cm}<{\centering}|m{0.5cm}<{\centering}}
    \hline\hline
         \multirow{2}{*}{Model} & \multicolumn{3}{c|}{mAP@IoU(\%)} &\multirow{2}{*}{Avg}  \\\cline{2-4}
         ~ & 0.5 &0.75 & 0.95 & ~  \\\hline\hline
         BaS-Net \cite{lee2019background} & 34.5 & 22.5 & 4.9 &	22.2 \\
         {UM \cite{lee2021weakly}} & {37.0}&	{23.9}&	{5.7}	&{23.7}\\
         Gong \emph{et al.} \cite{gongself} & 41.8 & 26.2 & 5.0 & 26.0 \\
         UGCT \cite{yang2021uncertainty} & 39.1 & 22.4 & 5.8 & 23.8\\
         FAC-Net \cite{huang2021foreground}& 37.6 & 23.9 & 5.7 & 23.7 \\
         FTCL \cite{gao2022fine} & 40.0 & 24.3 & \textbf{6.4} & 24.8\\
         ASM-Loc \cite{he2022asm} &41.0 & 24.9 & 6.2 & 25.1\\
         Huang \emph{et al.} \cite{huang2022weakly} & 40.6 & 24.6 & 5.9 & 25.0 \\
         Ours & \textbf{42.3} & \textbf{26.4} & 6.1 & \textbf{26.4}\\
         \hline\hline
    \end{tabular}
    \end{center}
    \label{tab:3}
\vspace{-0.75cm}
\end{table}

\subsubsection{Evaluation Metrics} We evaluate the proposed method for action localization by using mean Average Precision (mAP). the prediction proposal is considered as correct if its action category is predicted correctly and overlaps significantly with the ground truth segment (based on the IoU threshold). The official evaluation code of ActivityNet is used for measurement.

\subsubsection{Implementation Details}
Following previous work \cite{paul2018w,narayan20193c,ding2021kfc,2020Adversarial}, the optical flow maps are generated by using the TV-L1 algorithm \cite{inbook} and we use I3D network \cite{DBLP:conf/cvpr/CarreiraZ17} pre-trained on the Kineics dataset \cite{article2017} to extract both RGB and optical flow features without fine-turning. Same as the previous method \cite{paul2018w,narayan20193c,li2021multi,2020Adversarial,hong2021cross}, we sample continuous non-overlapping 16 frames from video as a snippet, where the features for each modal of each segment are 1024-dimension. 

The proposed method is implemented in PyTorch 1.8.0. Then during the training stage, we sample 320 segments for each video in THUMOS14 as \cite{paul2018w,narayan20193c} and 60 segments for each video in ActivityNet as \cite{hong2021cross}, while all segments are taken during the test stage.  In training, we set the batch size to 10 for THUMOS14 and ActivityNet. In addition, we use Adam \cite{kingma2014adam} with a learning rate of 0.0005 and weight decay of 0.001 to optimize our model for about 25,000 iterations. For $\alpha$, we set it as 0.25 to impose a consistency constraint. Besides, we set $\gamma = 0.6$ in Section \ref{subsec:AIC }  to collect pseudo-action segments and pseudo temporal context segments. {For the top-k multiple-instance learning loss, the number of selected top-k segments in a video $k$ is set as T/8, where T is the number of segments included in the video.} In addition, the training process requires two iterations to obtain more accurate pseudo-labels. All experiments are run on a Nvidia TITAN X GPU.

\subsection{Comparison with State-of-the-Art}
\label{subsubsec: SOTA}
We compare the proposed method with current weakly supervised state-of-the-art methods in this section. The result are shown in Tables \ref{tab:1}, and \ref{tab:3}. For THUMOS14 dataset, compared with current WTAL methods such as \cite{hong2021cross,huang2021foreground,huang2022multi,yang2021acgnet,huang2022weakly}, the proposed method significantly surpass them at almost all IoUs. Especially at average mAP(0.1:0.5) and average mAP(0.1:0.7), we surpass the state-of-the-art method by 1.1\% and 1.3\%. We can also note that the proposed method outperforms even some Weak+ methods at low IoU thresholds. Even if fully supervised methods utilize frame-level supervisions, indicating the effectiveness of our method. On the larger ActivityNet1.3 dataset, consistent with the results on THUMOS14, the proposed method still outperforms current work on most metrics.  

\subsection{Ablation Studies}
\label{subsubsec: ablation}
\subsubsection{Iterative strategy is necessary} 
\label{subsubsc:Iter}
In Section \ref{subsec:AIC }, the pseudo-labels generated by the pre-trained model are used to collect pseudo-action segment features and pseudo temporal context segment features in videos. To get more accurate pseudo labels, this paper adopts an iterative strategy. Concretely, first in iteration 1, a pre-trained baseline model without bidirectional semantic consistency loss is used to generate pseudo-labels to guide our model. Then in iteration 2, we update the pre-trained model parameters with the student network parameters of the original branch in the model from iteration 1. Similarly, the student network of the original branch trained in the $itr$-th iteration will be used as the pre-trained model for the $itr+1$-th iteration. In this paper, $itr$ is eventually set to 3 {since the performance saturates after that.} The experimental results for different iterations are shown in Table \ref{tab:4}, which verifies the effectiveness of the iterative strategy. {In addition, we added another experiment that using GT frame-level labels instead of pseudo labels in the model in the last row of the Table \ref{tab:4}. In this paper, 
we measure the quality of pseudo-labels by the Precision metric $q$ in the training set, i.e. the proportion of real foreground segments in the segments detected as foreground in the pseudo-labels. The results show that high-quality pseudo-labels will lead to more accurate action detection results.}

\begin{figure}[!htbp]
\begin{center}
    \includegraphics[width=1\linewidth]{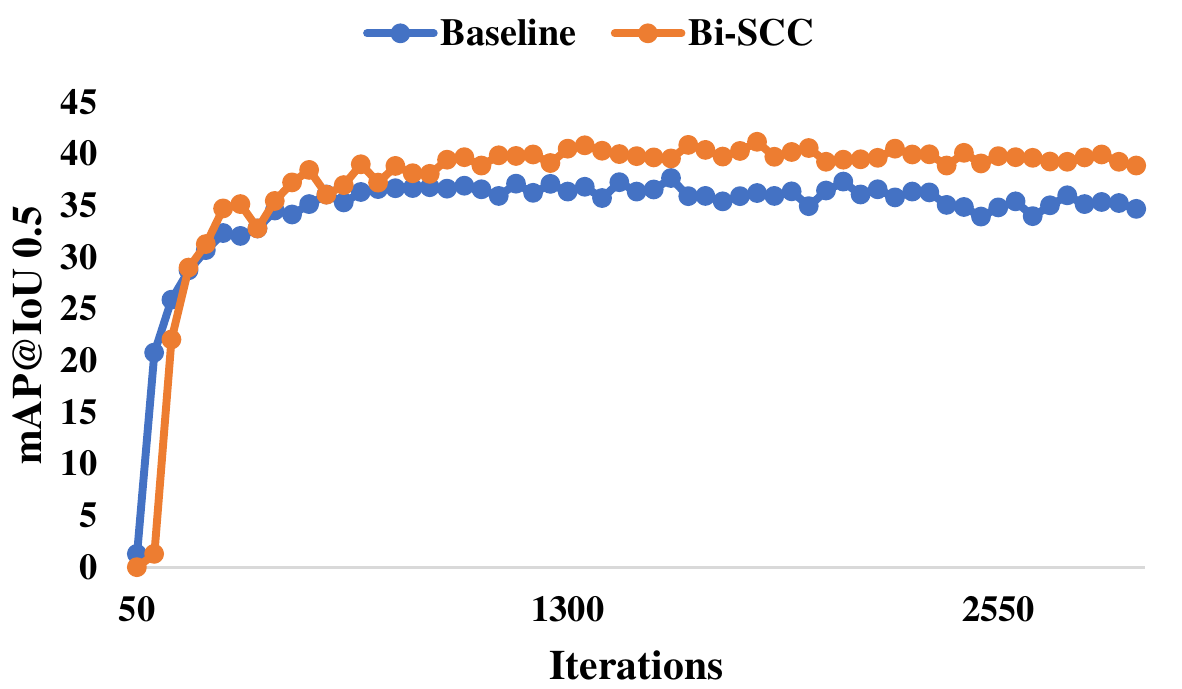}
\end{center}
   \caption{{Comparison of the accuracy curve of baseline model and the final Bi-SCC.}}
\label{fig:tra}
\vspace{-0.5cm}
\end{figure}

\begin{table}[!tbp]
    \centering
    \setlength{\abovecaptionskip}{-0.1cm}
    \caption{Experimental results of different iterations in THUMOS14 dataset, where `baseline' represents the baseline model used as the pre-training model for generating pseudo-labels in the first iteration {,`GT' means to use the ground truth frame-level annotations instead of pseudo-labels in the model and `$q$' is used to measure the quality of pseudo-labels.}}
    \begin{center}
    \begin{tabular}{m{3cm}<{\centering}|m{0.5cm}<{\centering}m{0.5cm}<{\centering}m{0.5cm}<{\centering}m{0.5cm}<{\centering}m{0.5cm}<{\centering}|m{0.5cm}<{\centering}}
    \hline\hline
         \multirow{2}{*}{{model}} & \multicolumn{5}{c}{mAP@IoU(\%)} &\multicolumn{1}{|c}{Avg}  \\\cline{2-6}
         ~ & 0.3 &0.4 & 0.5 & 0.6 & 0.7& 0.1:0.7  \\\hline\hline
        baseline ({w/o pseudo}) & 54.38 & 45.43 & 37.47 &25.29 & 13.25 & 44.25\\
        {Iteration 1 ($q$=0.552)} & 55.92 & 46.49 & 39.11 & 26.64 & 13.55 & 45.43 \\
        {Iteration 2 ($q$=0.601)} & 56.58 & 47.57 & 39.76 & 27.23 &13.42 & 45.97 \\
        {Iteration 3 ($q$=0.633)} & \textbf{57.24} & \textbf{47.98} & \textbf{40.44}  & 27.49 &14.39 & \textbf{46.60} \\
        {Iteration 4 ($q$=0.632)} 
        &{57.24}   &{47.85} 
        &{40.42} 
        &{\textbf{27.54}}  &{\textbf{14.55}} 
        &{46.56} \\
        \hline
        {GT($q$=1.0)}
        &{57.86}
        &{49.02}
        &{41.81}
        &{28.66}
        &{15.15}
        &{47.42} \\
         \hline\hline
    \end{tabular}
    \end{center}
    \label{tab:4}
\vspace{-0.5cm}
\end{table}
\subsubsection{{Effectiveness of each component.}}
\label{subsec:ASBiSCC}
{To validate the effectiveness of each component in the proposed method, we designed a series of comparative experiments as shown in Table \ref{tab:5}.
We compare the proposed Bi-SCC model with six simplified versions in the first six rows of Table \ref{tab:5}: 1) the baseline model; 2) the baseline model with $Inter$-TCA; 3) the baseline model with $Intra$-TCA; 4) the baseline model with both $Intra$- and $inter$-TCA; 5) The Bi-SCC model without $Intra$-TCA, and 6) The Bi-SCC model without $Inter$-TCA.}

\begin{table}[!tbp]
    \centering
    \setlength{\abovecaptionskip}{-0.1cm}
    \caption{Results of ablation studies on the THUMOS14 dataset, {where `Bi-SCC' means that we improve the semantic consistency constraint in a bidirectional manner.}}
    \begin{center}
    \begin{tabular}{m{0.55cm}<{\centering}m{0.55cm}<{\centering}|m{0.45cm}<{\centering}|m{0.42cm}<{\centering}m{0.42cm}<{\centering}m{0.42cm}<{\centering}m{0.42cm}<{\centering}m{0.42cm}<{\centering}|m{0.3cm}<{\centering}}
    \hline\hline
         \multicolumn{3}{c|}{{Method}} & 
         \multicolumn{5}{c|}{mAP@IoU(\%)} &
         \multicolumn{1}{c}{Avg}   \\\cline{1-9}
         \multicolumn{1}{c|}{$Inter$-} & \multicolumn{1}{c|}{$Intra$-} & \multicolumn{1}{c|}{Bi-} &
         {} & {} & {}& {}&{} &{}\\
         \multicolumn{1}{c|}{TCA} & \multicolumn{1}{c|}{TCA} &
         \multicolumn{1}{c|}{SCC}&
         \multirow{-2}{*}{0.3} & \multirow{-2}{*}{0.4} &\multirow{-2}{*}{0.5} & \multirow{-2}{*}{0.6} & \multirow{-2}{*}{0.7} & \multicolumn{1}{c}{\multirow{-2}{*}{{0.1:0.7}}} \\ \hline\hline
$\times$      &$\times$     & $\times$  & 54.38 & 45.43& 37.44& 25.22 & 13.25& \multicolumn{1}{l}{44.24}       \\
\checkmark      & $\times$     & $\times$     & 56.37& 46.86& 39.27& 26.59& 13.58& \multicolumn{1}{l}{45.56}       \\
$\times$      & \checkmark     & $\times$    & 55.26& 45.71& 38.32& 25.53& 13.42& \multicolumn{1}{l}{44.69}       \\
\checkmark     & \checkmark     & $\times$     & 56.94& 47.04& 39.62& 26.99& 13.39& \multicolumn{1}{l}{45.91}      \\
\checkmark      & $\times$     & \checkmark     & 57.04& 47.22& 40.02& 27.12& 14.17& \multicolumn{1}{l}{46.16}     \\
$\times$      & \checkmark    & \checkmark     & 56.18& 46.53& 38.46& 25.51& 12.89& \multicolumn{1}{l}{45.14}       \\
\checkmark      & \checkmark     & \checkmark     & \textbf{57.24}& \textbf{47.98}& \textbf{40.44}& \textbf{27.49}& \textbf{14.39}& \multicolumn{1}{l}{\textbf{46.60}}   \\   
         \hline\hline
    \end{tabular}
    \end{center}
    \label{tab:5}
\vspace{-0.5cm}
\end{table}

\begin{table}[!htbp]
    \centering
    \setlength{\abovecaptionskip}{-0.1cm}
    \caption{Results of ablation studies on the ActivityNet1.3 dataset, {where `Bi-SCC' means that we improve the semantic consistency constraint in a bidirectional manner.}}
    \begin{center}
    \begin{tabular}{m{0.55cm}<{\centering}m{0.55cm}<{\centering}|m{0.65cm}<{\centering}|m{0.5cm}<{\centering}m{0.5cm}<{\centering}m{0.5cm}<{\centering}|c}
    \hline\hline
         \multicolumn{3}{c|}{{Method}} & 
         \multicolumn{3}{c|}{{mAP@IoU(\%)}} &
         \multicolumn{1}{c}{{Avg}}   \\\cline{1-7}
         \multicolumn{1}{c|}{{$Inter$-}} & \multicolumn{1}{c|}{{$Intra$-}} & \multicolumn{1}{c|}{{Bi-SCC}} &
         {} & {} & {}& {}\\
         \multicolumn{1}{c|}{{TCA}} & \multicolumn{1}{c|}{{TCA}} &
         {}&
         \multirow{-2}{*}{{0.5}} & \multirow{-2}{*}{{0.75}} & \multirow{-2}{*}{{0.95}} &  \multicolumn{1}{c}{\multirow{-2}{*}{{0.5:0.95}}} \\ \hline\hline
{$\times$}      &{$\times$}     &{$\times$} 
& {40.91} & {25.36}& {5.78}&  \multicolumn{1}{l}{{25.49}}       \\
{\checkmark}& {$\times$}& {$\times$}
& {41.68}& {25.62}& {5.95}&  \multicolumn{1}{l}{{25.94}}       \\
{$\times$}      & {\checkmark}      & {$\times$}   
& {41.39}& {25.45}& {6.07}&  \multicolumn{1}{l}{{25.83}}       \\
{\checkmark} 
& {\checkmark} 
& {$\times$}      
& {41.97}& {26.23}& {6.04}&  \multicolumn{1}{l}{{26.04}}      \\
{\checkmark}      & {$\times$}     & {\checkmark} 
& {42.09}& {26.22}& {5.99}&  \multicolumn{1}{l}{{26.24}}     \\
{$\times$}      & {\checkmark}    & {\checkmark}      
& {41.77}& {25.98}& {6.04}&  \multicolumn{1}{l}{{26.09}}       \\
{\checkmark}      & {\checkmark}      & {\checkmark}      & \textbf{{42.28}}& \textbf{{26.44}}& \textbf{{6.09}}&  \multicolumn{1}{l}{\textbf{{26.44}}}   \\   
         \hline\hline
    \end{tabular}
    \end{center}
    \label{tab:51}
\vspace{-0.5cm}
\end{table}

{
Specifically, different from the final Bi-SCC model where each of the two branches contains a teacher-student network, the structure of the baseline model only contains a teacher-student network. In the baseline model, the original video features are directly input into the student network, i.e. a T-CAM generation module described in Section \ref{subsec:T-CAM Generation} to generate T-CAM $\mathbf{S}$ and $\mathbf{\bar{S}}$ according to Eq. \ref{E4} and Eq \ref{E5}. Then the student network can be optimized by Eq. \ref{E10}-Eq. \ref{E14}.}

{ 
At the same time, in the baseline model, the original video features without any data augmentation can be also input to the teacher network whose parameters are updated by the parameters of the student network by the exponential moving average (EMA) strategy. Then the teacher network will output another background-suppressed T-CAM $\mathbf{\bar{S}}^{\mathcal{T}}$, and constrain the output of the student network through Eq. \ref{E7}. The weakly-supervised training procedure for the baseline model
can be also shown in Algorithm \ref{A:0}.}
{In addition, it is worth noticing that the simplified baseline model we mentioned here is exactly the pre-training baseline used to generate pseudo labels in Section \ref{subsec:AIC }}.

As shown in the first four rows of Table \ref{tab:5}, we can find that when using $Inter$-TCA and $Intra$-TCA temporal context augmentation for semantic consistency constraint, the performance of the model is significantly improved by 1.3\% and 0.5\% in terms of, respectively. When both augmentation schemes are used simultaneously, average mAP will increase by 1.7\%. We also compare the performance differences of the models with and without the bidirectional supervision shown in the last four rows of Table \ref{tab:5}, the bidirectional semantic consistency constraint brings another {0.7\%} improvement to the model without that on the average mAP, justifying the effectiveness of the bidirectional manner. {Apart from THUMOS14, we also conduct ablation experiments on the ActivityNet1.3 dataset, as shown in the Table \ref{tab:51}. The two temporal context augmentation strategies and SCC can bring 0.55\% improvement compared to the baseline model on the average mAP. And the use of Bi-SCC also further brings a 0.4\% improvement. The experimental results also verify the effectiveness of the proposed method.}

\begin{table}[!htbp]
    \centering
    \setlength{\abovecaptionskip}{-0.1cm}
    \caption{{Experimental results on localization performance of different network structures in THUMOS14 dataset.}}
    \begin{center}
    \begin{tabular}{m{3cm}<{\centering}|m{0.4cm}<{\centering}m{0.4cm}<{\centering}m{0.4cm}<{\centering}m{0.4cm}<{\centering}m{0.5cm}<{\centering}|m{0.5cm}<{\centering}}
    \hline\hline
         \multirow{2}{*}{Network} & \multicolumn{5}{c}{mAP@IoU(\%)} &\multicolumn{1}{|c}{Avg}  \\\cline{2-6}
         ~ & 0.3 &0.4 & 0.5 & 0.6 & 0.7& 0.1:0.7  \\\hline\hline
        baseline & 54.38 & 45.43 & 37.47 &25.29 & 13.25 & 44.25\\
        {SCC (w/o CTG)} &  56.94& 47.04 &39.62 &26.99& 13.39 &45.91\\
        {SCC v2 (w/o CTG)} & 55.10 & 46.09 & 38.29 & 25.51 & 12.65 & 44.58\\
        {SCC (with CTG)}
        &{57.21}
        &{47.58}
        &{40.26}
        &{27.30}
        &{14.33}
        &{46.29}\\
        {Bi-SCC (w/o CTG)}
        &{57.01}
        &{47.65}
        &{40.25}
        &{27.42}
        &{14.23}
        &{46.34}\\
        {Bi-SCC (with CTG)} &  \textbf{57.24}& \textbf{47.98}& \textbf{40.44}& \textbf{27.49}& \textbf{14.39}& \textbf{46.60} \\
        
        {Bi-SCC v2 (with CTG)} & 56.21 & 46.87 & 39.42 & 26.24 & 13.82 & 45.63 \\
        
        {Bi-SCC (with CTG) with intra-video consistency}
        &{56.23}
        &{47.30}
        &{40.09}
        &{26.81}
        &{14.25}
        &{45.87}\\
        
         \hline\hline
    \end{tabular}
    \end{center}
    \label{tab:6}
\vspace{-0.5cm}
\end{table}

{The comparison of accuracy curves between the baseline model and the final Bi-SCC is presented in Figure \ref{fig:tra}. With the increase in training time, the performances of baseline and Bi-SCC models are constantly improving. The baseline model converges after 900 iterations and mAP@IoU 0.5 reaches 37.4\%, while the Bi-SCC converges after 1500 iterations and mAP@IoU 0.5 reaches 40.4\%, which surpasses the baseline model by 3\%. In addition, due to the more complex structure, the training time of the Bi-SCC model is almost twice as long as that of the baseline model.
}

\begin{table}[!htbp]
    \centering
    \setlength{\abovecaptionskip}{-0.1cm}
    \caption{Ablation studies of the comprehensive T-CAM generation module in THUMOS14 dataset. }
    \begin{center}
    \begin{tabular}{m{2cm}<{\centering}|m{0.5cm}<{\centering}m{0.5cm}<{\centering}m{0.5cm}<{\centering}m{0.5cm}<{\centering}m{0.5cm}<{\centering}|m{0.5cm}<{\centering}}
    \hline\hline
         \multirow{2}{*}{Module} & \multicolumn{5}{c}{mAP@IoU(\%)} &\multicolumn{1}{|c}{Avg}  \\\cline{2-6}
         ~ & 0.3 &0.4 & 0.5 & 0.6 & 0.7& 0.1:0.7  \\\hline\hline
        w/o &56.75& 47.54 &39.93 &26.92& 13.4 &46.09\\
        avg &   \textbf{57.56}& 47.94 &40.14 &27.33& 14.01 &46.39\\
        max &  57.24& \textbf{47.98}& \textbf{40.44}& \textbf{27.49}& \textbf{14.39}& \textbf{46.60} \\
         \hline\hline
    \end{tabular}
    \end{center}
    \label{tab:ct}
\vspace{-0.5cm}
\end{table}

{In addition, we compare the applied framework against other simplified versions in Table \ref{tab:6}, where we rename the ``baseline with both $Intra$- and $Inter$-TCA'' as ``SCC''.} Besides, in the SCC v2 model in Table \ref{tab:6}, we exchange the inputs of the teacher network and the student network in the SCC model to evaluate the quality of the constructed augmented video. Similarly, in Bi-SCC v2 model, we test using the output of the student network of the augmentation branch to evaluate the quality of the constructed augmented video. The experimental results in Table \ref{tab:6} shows that the model with the augmented video as input still achieve good localization results, verifying the proposed temporal context augmentation strategy effectively preserves the positive actions contained in the video.

\begin{table}[!tbp]
    \centering
    \setlength{\abovecaptionskip}{-0.1cm}
    \caption{Experimental results of different semantic consistency constraint function in THUMOS14 dataset.}
    \begin{center}
    \begin{tabular}{m{1cm}<{\centering}|m{0.5cm}<{\centering}m{0.5cm}<{\centering}m{0.5cm}<{\centering}m{0.5cm}<{\centering}m{0.5cm}<{\centering}|m{0.5cm}<{\centering}}
    \hline\hline
         \multirow{2}{*}{Method} & \multicolumn{5}{c}{mAP@IoU(\%)} &\multicolumn{1}{|c}{Avg}  \\\cline{2-6}
         ~ & 0.3 &0.4 & 0.5 & 0.6 & 0.7& 0.1:0.7  \\\hline\hline
        MSE & 55.88 & 46.35 & 39.66 &27.00 & 13.73& 45.49\\
        L1 & 55.21 & 46.09 & 39.11 & 26.45 & 13.57 & 44.87 \\
        KL & \textbf{57.24}& \textbf{47.98}& \textbf{40.44}& \textbf{27.49}& \textbf{14.39}& \textbf{46.60} \\
         \hline\hline
    \end{tabular}
    \end{center}
    \label{tab:7}
\vspace{-0.5cm}
\end{table}

{Table \ref{tab:6} and Figure \ref{fig:Qua} shows that compared with the SCC model, the Bi-SCC model can effectively improve the completeness of the prediction of the positive action and get higher localization accuracy, verifying the effectiveness of the cross-supervision in Eq. \ref{E8}. Furthermore, in Table \ref{tab:6}, we also verify the effectiveness of the proposed CTG module through ablation experiments. 
Compared with the model without CTG model, both SCC model and Bi-SCC model with CTG achieved higher localization accuracy, verifying the effectiveness of the proposed CTG module.}
\begin{table}[!htbp]
    \centering
    \setlength{\abovecaptionskip}{-0.1cm}
    \caption{Experimental results of different $K$ in THUMOS14 dataset.}
    \begin{center}
    \begin{tabular}{m{1cm}<{\centering}|m{0.5cm}<{\centering}m{0.5cm}<{\centering}m{0.5cm}<{\centering}m{0.5cm}<{\centering}m{0.5cm}<{\centering}|m{0.5cm}<{\centering}}
    \hline\hline
         \multirow{2}{*}{$K$} & \multicolumn{5}{c}{mAP@IoU(\%)} &\multicolumn{1}{|c}{Avg}  \\\cline{2-6}
         ~ & 0.3 &0.4 & 0.5 & 0.6 & 0.7& 0.1:0.7  \\\hline\hline
        1 & 57.19 & 47.21 & 40.09 & 27.18 & 14.06 & 46.16 \\
        2 & \textbf{57.49} & 47.45 & 40.24 & 27.23 &14.22 & 46.32 \\
        3 & 57.24 & \textbf{47.98} & \textbf{40.44}  & \textbf{27.49} &\textbf{14.39} & \textbf{46.60} \\
        4 & 56.55 & 46.98 & 40.13 & 26.77 &13.78 & 46.13 \\
         \hline\hline
    \end{tabular}
    \end{center}
    \label{tab:K}
\vspace{-0.5cm}
\end{table}
{Finally, in the last row of Table \ref{tab:6}, we also show the performance of the network trained with intra-video consistency between $\bar S$ and ${\bar S}^{C\mathcal{T}}$. 
At this point, the T-CAM $\bar S$ is constrained by ${\bar S}^{C\mathcal{T}}$ from the original branch and ${\bar S}^{C\mathcal{T}’}$ from the augmentation branch at the same time.  Simultaneously supervising the same target by two different pseudo-labels causes confusion and reduces the experimental effect.}

Besides, we also evaluate the effectiveness of the proposed comprehensive T-CAM generation (CTG) module in Table \ref{tab:ct}. The model with the proposed CTG module can outperform that without the CTG module, indicating the necessity for the comprehensive T-CAM strategy.

\subsubsection{Ablation studies of different types of semantic consistency constraint}
We evaluate the effect of the different types of the semantic consistency constraint and report the results in Table \ref{tab:7}. The results show that almost all types of semantic consistency constraint functions can outperform the current state-of-the-art results shown in Table \ref{tab:1}, indicating the effectiveness of the proposed semantic consistency constraint. Besides, the experimental results in Table \ref{tab:7} show that the KL divergence loss is more suitable for the proposed semantic consistency constraint.

\begin{table}[!htbp]
    \centering
    \setlength{\abovecaptionskip}{-0.1cm}
    \caption{{Experimental results of different data augmentation in CTG module in THUMOS14 dataset.}}
    \begin{center}
    \begin{tabular}{m{3cm}<{\centering}|m{0.4cm}<{\centering}m{0.4cm}<{\centering}m{0.4cm}<{\centering}m{0.4cm}<{\centering}m{0.5cm}<{\centering}|m{0.5cm}<{\centering}}
    \hline\hline
         \multirow{2}{*}{{Method}} & \multicolumn{5}{c}{{mAP@IoU(\%)}} &\multicolumn{1}{|c}{{Avg}}  \\\cline{2-6}
         ~ & {0.3} &{0.4} & {0.5} & {0.6} & {0.7}& {0.1:0.7}  \\\hline\hline
{w/o data augmentation} & {57.04} 
& {47.22} 
& {40.02} 
& {27.12}
& {14.17} 
& {46.16} \\
{temporal filp} & {56.72} &{47.37} &{39.95} &{27.19} &{14.19} &{46.20}\\
{temporal resolution transformation} & {56.83} 
& {47.38} 
& {39.95} 
& {27.56}
& {14.22} 
& {46.18} \\
{random mask} 
& {56.27} 
& {47.59} 
& {40.32} 
&{27.28}  
& {14.12} 
& {46.24} \\
{Intra-TCA}
& {\textbf{57.24}}
& \textbf{47.99}
& \textbf{40.44} 
& \textbf{27.49}  
& \textbf{14.39}
& \textbf{46.60} \\
\hline
{temporal filp + temporal resolution transformation} 
& {56.32} 
& {48.15}
& {40.65}
&{27.31}  
& {14.17} 
& {46.27} \\
{random mask + temporal resolution transformation} & {56.75} &{48.09} &{40.68} &{27.86} &{14.60} &{46.58}\\

{Intra-TCA + temporal resolution transformation} 
&\textbf{{57.26}} 
&\textbf{{48.82}} 
&\textbf{{41.56}} 
&\textbf{{28.08}} &\textbf{{14.78}} &\textbf{{46.78}}\\
\hline\hline
    \end{tabular}
    \end{center}
    \label{tab:K2}
\vspace{-0.5cm}
\end{table}
\subsubsection{Relative weights on loss functions} In our work, we investigated the effect of hyper-parameter $\alpha$ on model performance. The hyper-parameter $\alpha$ controls the relative weights of the bidirectional semantic consistency constraint. As shown in Figure \ref{fig:hyper}, the proposed method performs best with $\alpha = 0.25$. That's because a high $\alpha$ will interfere with the training of the respective classifiers for each branch, while a too low  $\alpha$ will cause the consistency constraint to be ignored. Moreover, the model without semantic consistency constraint will decrease by 2.7\% in average mAP. This shows that the semantic consistency constraint introduced in this work has a major effect on the better performance of our framework.

\subsubsection{Sensitivity to the threshold of the pseudo positive action segments}
\begin{figure}[!t]
\begin{center}
    \includegraphics[width=1\linewidth]{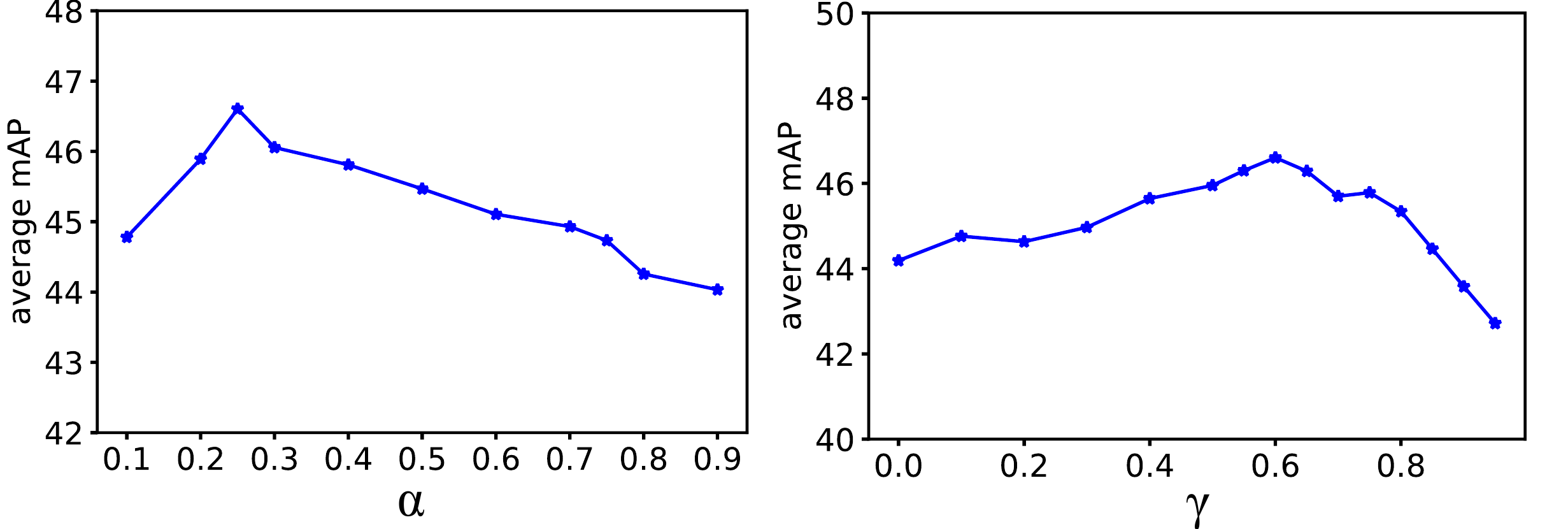}
\end{center}
   \caption{{(a) presents the variations in localization performance on THUMOS14 by changing the weights on the semantic consistency constraint. (b) presents the variations in localization performance on THUMOS14 by changing the threshold of the pseudo-action segments.}}
\label{fig:hyper}
\vspace{-0.5cm}
\end{figure}

We also investigated the effect of hyper-parameter $\gamma$ on model performance. The experimental results are shown in Fig \ref{fig:hyper},  the proposed method performs best with $\gamma = 0.6$.
Due to the co-scene actions in temporal context segments are easily confused as pseudo positive actions, the value of parameter $\gamma$ should not be too low. Besides, when the parameter gamma value is too high, the selected pseudo positive action segments would be too few and enabling the model over-focus on the most discriminative parts of the video, causing incomplete localization. Both of these cases would affect the performance of the proposed semantic consistency constraint.

\subsubsection{Sensitivity to the number of different augmentations applied in the CTG module}
In our work, we investigated the effect of hyper-parameter $K$ on model performance. As shown in Table \ref{tab:K}, the proposed method performs best with $K = 3$. Compared with the model without comprehensive T-CAM (\emph{i.e.} $K=1$), the proposed strategy effectively improves the localization performance. For higher K, the performance of the model will be degraded. We think this may be caused by the fact that the integrated T-CAM not only covers more action areas than a single T-CAM but also covers more background noise areas.

\begin{table}[!htbp]
    \centering
    \setlength{\abovecaptionskip}{-0.1cm}
    \caption{{Experimental results of different $k$ in the top-k multiple-instance learning loss in THUMOS14 dataset, where T is number of segments in a video.}}
    \begin{center}
    \begin{tabular}{m{1.5cm}<{\centering}|m{0.5cm}<{\centering}m{0.5cm}<{\centering}m{0.5cm}<{\centering}m{0.5cm}<{\centering}m{0.5cm}<{\centering}|m{0.5cm}<{\centering}}
    \hline\hline
         \multirow{2}{*}{{$k$}} & \multicolumn{5}{c}{{mAP@IoU(\%)}} &\multicolumn{1}{|c}{{Avg}}  \\\cline{2-6}
         ~ & {0.3} &{0.4} & {0.5} & {0.6} & {0.7}& {0.1:0.7}  \\\hline\hline

{T/2} 
& {53.44} 
& {44.63} 
& {36.78} 
&{24.75}
& {12.96} 
& {43.45} \\
{T/4} 
& {55.28} 
& {47.10} 
& {39.09} 
&{26.54}  
& {14.31} 
& {45.37} \\
{T/8} 
& {\textbf{57.24}}
& {\textbf{47.98}}
& {\textbf{40.45}}
&{\textbf{27.49}}  
& {\textbf{14.39}} 
& {\textbf{46.60}} \\

{T/12}
& {56.17}
& {47.45}
& {39.79} 
&{26.87}  
& {13.98}
& {45.67} \\
\hline\hline
    \end{tabular}
    \end{center}
    \label{tab:K3}
\vspace{-0.5cm}
\end{table}

{Furthermore, considering the necessity of utilizing intra-video data augmentation strategies in the CTG module to help generate a complete T-CAM, we have added a comparative experiment between the $Intra$-TCA strategy used in this paper and other data augmentation strategies in the CTG module.

{
Table \ref{tab:K2} shows the proposed Intra-TCA strategy achieves better results than other data augmentation strategies when we use only one data augmentation method in the CTG module. Specifically, compared with other data augmentation methods such as “temporal flip”, “temporal resolution transformation” and “random mask”, the proposed “$Intra$-TCA” surpasses them by 0.40\%, 0.42\% and 0.36\% on average mAP respectively.
}

{In addition, we also compared the effects of two data augmentation methods on model performance in the CTG module \ref{tab:K2}. We can find that applying both `$Intra$-TCA’ and `temporal resolution transformation’ can further achieve better performance than other methods. Specifically, the ``$Intra$-TCA + temporal resolution transformation'' is higher than ``temporal flip + temporal resolution transformation'' and ``random mask+ temporal resolution transformation'' 0.51\% and 0.30\% on average mAP.}

That may be because the proposed strategy breaks the correlation between actions and their inherent pre- and post-temporal contexts, leading to richer intra-video variation.}

\subsubsection{{Sensitivity to the number of selected segments in the top-k multiple-instance learning loss}}
{In this paper, we also analyze the influence of the number of selected top-k segments in each video $k$ on the experimental results in computing the top-k multi-instance learning classification loss. The experimental results are shown in the Table \ref{tab:K3}. As the number of selected top-k segments increases or decreases, the performance degrades. The reason is that at on the one hand, a higher number of $k$, noisy segments can corrupt the classification score, on the other hand, a lower $k$ may cause the model to ignore some action segments and cause the model to miss some short actions.}

\begin{table}[!tbp]
    \centering
    \setlength{\abovecaptionskip}{-0.1cm}
    \caption{Experimental results of different data augmentation strategies in THUMOS14 dataset.}
    \begin{center}
    \begin{tabular}{m{2.5cm}<{\centering}|m{0.5cm}<{\centering}m{0.5cm}<{\centering}m{0.5cm}<{\centering}m{0.5cm}<{\centering}m{0.5cm}<{\centering}|m{0.5cm}<{\centering}}
    \hline\hline
         \multirow{2}{*}{Method} & \multicolumn{5}{c}{mAP@IoU(\%)} &\multicolumn{1}{|c}{Avg}  \\\cline{2-6}
         ~ & 0.3 &0.4 & 0.5 & 0.6 & 0.7& 0.1:0.7  \\\hline\hline
        w/o & 54.38 & 45.43& 37.44& 25.22 & 13.25& 44.24\\
        gaussian noise & 54.62 & 45.73 & 37.97 & 25.73 & 13.19 & 44.69\\
        random mask & 52.65 & 44.04 & 36.85 &25.03 & 13.03& 43.41\\
        temporal resolution transformation & 55.91 & 45.70 & 38.62 & 26.07 & 13.35 & 44.70 \\
        random cut-paste & 54.26& 45.28 & 37.09 & 25.59& 13.34& 44.15\\
        our $Intra$-TCA & 55.26 & 45.71 & 38.32 & 25.53 & 13.42 & 44.69 \\
        our $Inter$-TCA & \textbf{56.37} & \textbf{46.86} & \textbf{39.27} & \textbf{26.59} & \textbf{13.58} & \textbf{45.57}\\
         \hline\hline
    \end{tabular}
    \end{center}
    \label{tab:8}
\vspace{-0.5cm}
\end{table}

\begin{table}[!bp]
    \centering
    \setlength{\abovecaptionskip}{-0.1cm}
    \caption{Experimental results of different data augmentation strategies in THUMOS14 dataset. `Bi-SCC' means the proposed semantic consistency constraint.}
    \begin{center}
    \begin{tabular}{m{4cm}<{\centering}|m{1.0cm}<{\centering}}
    \hline\hline
    \multirow{2}{*}{{Method}} &\multicolumn{1}{c}{{average mAP (\%)}} \\ 
    ~ &  {0.1:0.7}  \\\hline\hline
        {BaS-Net} \cite{lee2019background} & {34.21} \\
        {BaS-Net \cite{lee2019background} + Bi-SCC} & {35.42} \\\hline
        {HAM-Net} \cite{islam2021hybrid} & {39.86} \\
        {HAM-Net \cite{islam2021hybrid} + Bi-SCC} & {40.65} \\\hline
        {CO2-Net} \cite{hong2021cross} & {44.57} \\
        {CO2-Net \cite{hong2021cross} + Bi-SCC} & {45.78} \\
         \hline\hline
    \end{tabular}
    \end{center}
    \label{tab:9}
\vspace{-0.5cm}
\end{table}

\subsection{Compared with Other Temporal Context Augmentation Strategy}
\label{subsubsec: COTCA}
We compare the proposed two temporal context augmentation schemes with some other data augmentation methods, including random mask, temporal resolution transformation \cite{su2021improving,gongself}, gaussian noise and random cut-paste. As shown in Table \ref{tab:8}, the proposed $Inter$-TCA temporal context augmentation strategy surpasses all other temporal context augmentation methods. That's because compared with traditional data augmentation, the proposed $Inter$-TCA strategy can break the correlation between action instances and contextual background, and the models will focus on the internal information of the action instances rather than the context information to predict the categories they belong to. On the other hand, compare with the random cut-paste strategy which also synthesizes augmented videos, the $Inter$-TCA strategy preserves the internal frame order of action instances, thus preserving action semantic information helps the model to understand actions more effectively. Besides, the $Intra$-TCA strategy can also improve the model performance to a certain extent, but it only acts inside the video and its temporal context augmentation ability is weaker than that of the $Inter$-TCA strategy.

\subsection{Applying the semantic consistency constraint Strategy to Existing Method}
\label{subsubsec: ASCEM}

The proposed semantic consistency constraint strategy can also be generalized to other existing WTAL methods. Here, we impose the semantic consistency constraint proposed in this paper on Bas-net \cite{lee2019background}, Ham-net \cite{islam2021hybrid} and CO2-net \cite{hong2021cross} in Table \ref{tab:9}. The experimental results in Table \ref{tab:9} show that the proposed semantic consistency constraint strategy can indeed be applied to existing WTAL methods and improve their performance, verifying the generalizability of the proposed strategy.

\subsection{Qualitative Results}

\begin{figure}[!htbp]
\begin{center}
    \includegraphics[width=1.0\linewidth]{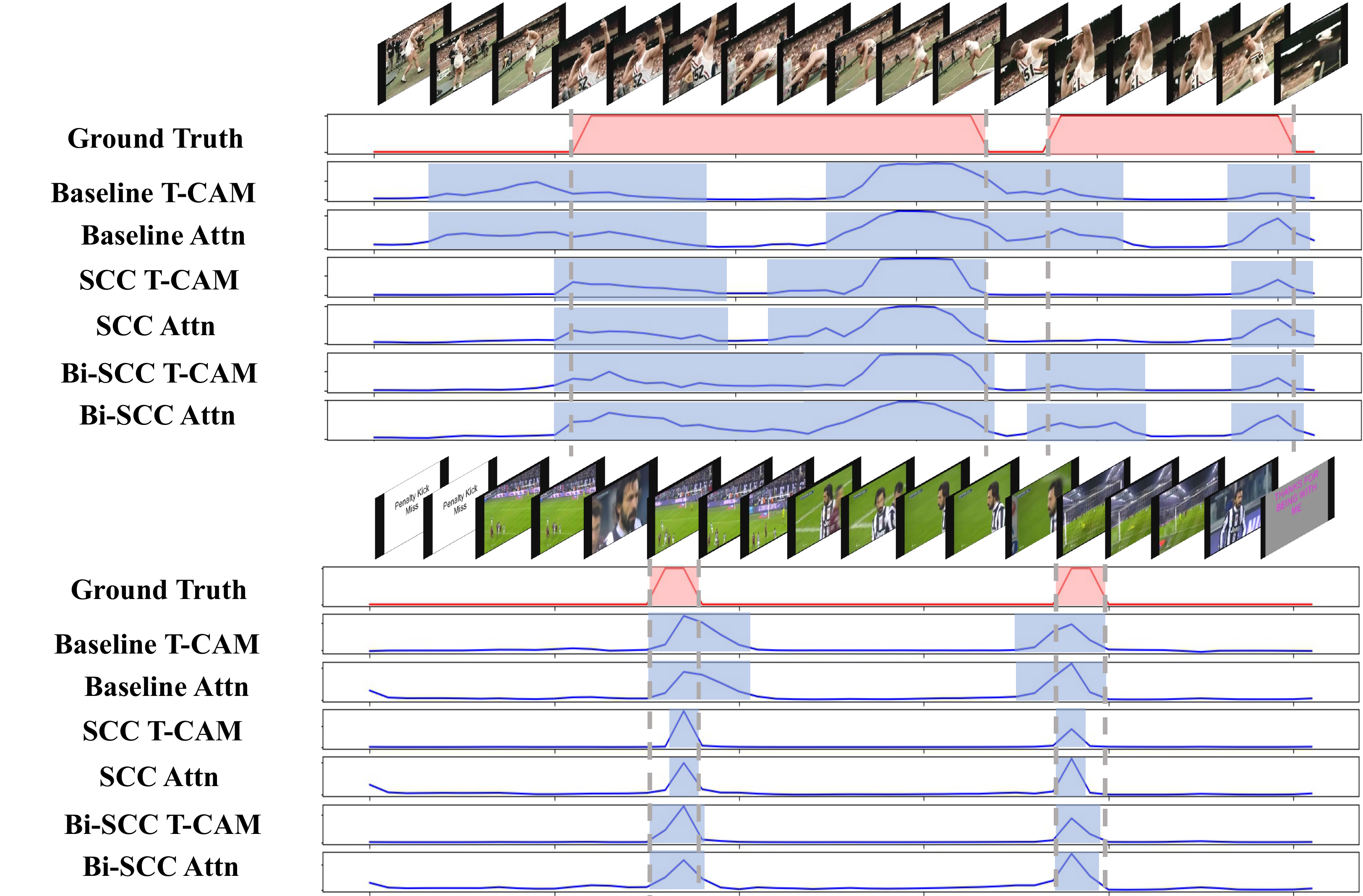}
\end{center}
   \caption{{Two prediction examples on THUMOS14 dataset. For each example, we show the input video, the ground truth, the T-CAM and attention generated by the baseline model, the T-CAM and attention generated by the model with the semantic consistency constraint, and the T-CAM and attention generated by the framework with the bidirectional semantic consistency constraint. }}
\label{fig:Qua}
\vspace{-0.5cm}
\end{figure}

Figure \ref{fig:Qua} shows the qualitative analysis of our method on some examples of the THUMOS14 \cite{THUMOS14} dataset. For each example, the top line represents the segment of the video, and the following five lines are in order the ground truth of the action in the video, the T-CAM and attention generated by the baseline model, the T-CAM and attention generated by the model with semantic consistency constraint. The last two lines of each example are the T-CAM and attention generated by the final Bi-SCC of this paper. It can be clearly noticed that our method locates the action instances more precisely and completely. The proposed method effectively alleviates the interference caused by the co-scene actions in temporal context, enabling the model to locate more precise action regions, verifying the effectiveness of the proposed temporal context augmentation and semantic consistency constraint. In addition, the bidirectional semantic consistency constraint strategy achieves an effective improvement in the performance of the model.

\label{subsubsec: Qualitative}
\section{Conclusion}
\label{sec:conclusion}
  
This paper explores the semantic consistency constraint for weakly supervised temporal action localization. Our method contains two temporal context augmentation strategies aiming to break the dependency between actions and co-occurrence temporal context. Besides, the semantic consistency constraint is used to ensure the generated T-CAMs of video action instances are invariant before and after temporal context augmentation. Furthermore, a comprehensive T-CAM knowledge distillation strategy and iteration strategy further improve the performance of the model. The proposed method outperforms all existing WTAL methods and achieves state-of-the-art performance on two challenging benchmarks and can consistently improve the performance of existing methods.


\ifCLASSOPTIONcaptionsoff
  \newpage
\fi
\bibliographystyle{IEEEtran}
\bibliography{./DCF}

\end{document}